\documentclass[journal]{IEEEtran}
\pdfoutput=1
\ifCLASSOPTIONcompsoc
  \usepackage[nocompress]{cite}
\else
  \usepackage{cite}
\fi

\ifCLASSINFOpdf

\else

\fi

\usepackage{times}
\usepackage{helvet}
\usepackage{courier}
\usepackage{amsmath}
\usepackage{amsthm}
\usepackage{multirow}
\usepackage{color}
\usepackage{algorithm}
\usepackage{algorithmic}
\usepackage{graphicx}
\usepackage{epstopdf}
\usepackage{caption}
\usepackage{float}
\DeclareCaptionType{copyrightbox}
\usepackage{subcaption}
\graphicspath{ {./figure/} }
\begin{document}

\title{Detecting Communities in Heterogeneous Multi-Relational Networks:
\\
A Message Passing based Approach}

\author{Maoying~Qiao,~
        Jun~Yu,~\IEEEmembership{IEEE Member,}
        Wei~Bian,~\IEEEmembership{IEEE Member,}
        and~Dacheng~Tao,~\IEEEmembership{IEEE Fellow}
\IEEEcompsocitemizethanks{\IEEEcompsocthanksitem M. Qiao is with Data61, CSIRO, Australia (email: maoying.qiao@csiro.au).
\IEEEcompsocthanksitem J. Yu is with the School of Computer Science, Hangzhou Dianzi University, China (email: yujun@hdu.edu.cn).
\IEEEcompsocthanksitem W. Bian is with the Centre for Artificial Intelligence, the University of Technology Sydney, Australia (email: wei.bian@uts.edu.au).
\IEEEcompsocthanksitem D. Tao is with the UBTECH Sydney Artificial Intelligence Centre and the School of Computer Science, in the Faculty of Engineering and Information Technologies, at the University of Sydney, 6 Cleveland St, Darlington, NSW 2008, Australia (email: dacheng.tao@sydney.edu.au).
}
}

\IEEEtitleabstractindextext{
\begin{abstract}
Community is a common characteristic of networks including social networks, biological networks, computer and information networks, to name a few.
Community detection is a basic step for exploring and analysing these network data.
Typically, homogenous network is a type of networks which consists of only one type of objects with one type of links connecting them.
There has been a large body of developments in models and algorithms to detect communities over it.
However, real-world networks naturally exhibit heterogeneous qualities appearing as multiple types of objects with multi-relational links connecting them.
Those heterogeneous information could facilitate the community detection for its constituent homogeneous networks, but has not been fully explored.
In this paper,
we exploit heterogeneous multi-relational networks (HMRNet) and propose an efficient message passing based algorithm to simultaneously detect communities for all homogeneous networks.
Specifically, an HMRNet is reorganized into a hierarchical structure with homogeneous networks as its layers and heterogeneous links connecting them.
To detect communities in such an HMRNet, the problem is formulated as a maximum a posterior (MAP) over a factor graph.
Finally a message passing based algorithm is derived to find a best solution of the MAP problem.
Evaluation on both synthetic and real-world networks
confirms the effectiveness of the proposed method.
\end{abstract}

}

\maketitle
\IEEEdisplaynontitleabstractindextext

\IEEEpeerreviewmaketitle

\vspace{3em}
\IEEEraisesectionheading{\section{Introduction}\label{sec:introduction}}

\IEEEPARstart{C}{luster}
structures, or communities, where the links within intra-community are dense and inter-community sparse, are one of the key characteristics for modern networks. For instance, circles of friends prominently appear in social networks; and functionally interacting proteins present different groups in protein-protein interaction networks. Finding such cluster structures, which is termed community detection, is a fundamental problem in network analysis.
It has received increasing attention from considerably wide disciplines, from sociology and biology to physics and computer science \cite{jia2012community,leskovec2008statistical,du2007community}.

\begin{figure*}[tbh]
\centering
\begin{subfigure}[b]{0.5\textwidth}
\centering
\includegraphics[height=0.5\textwidth]{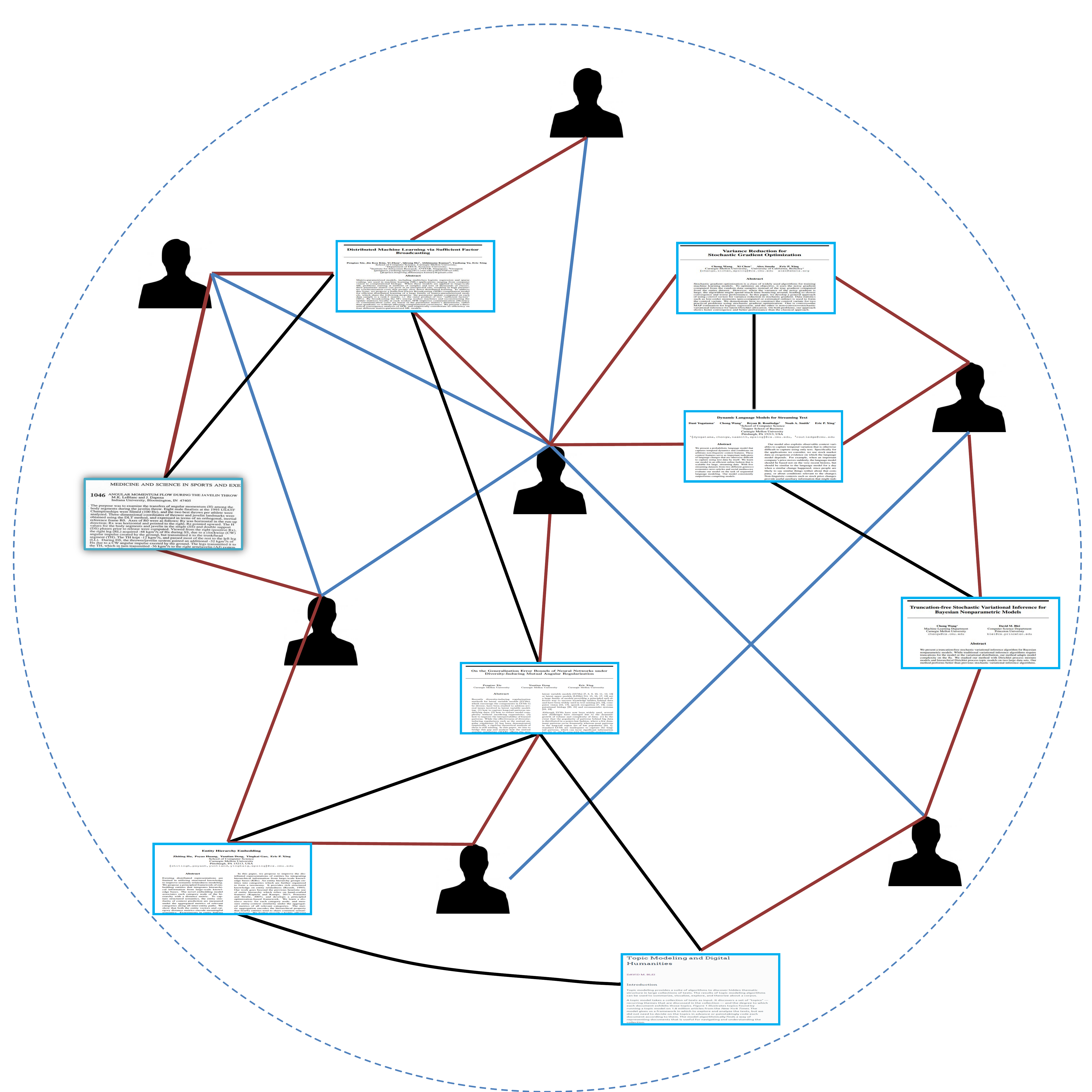}
\caption{An HMRN in the wild}
\label{fig:HINexample1}
\end{subfigure}
\begin{subfigure}[b]{0.5\textwidth}
\centering
\includegraphics[height=0.5\textwidth]{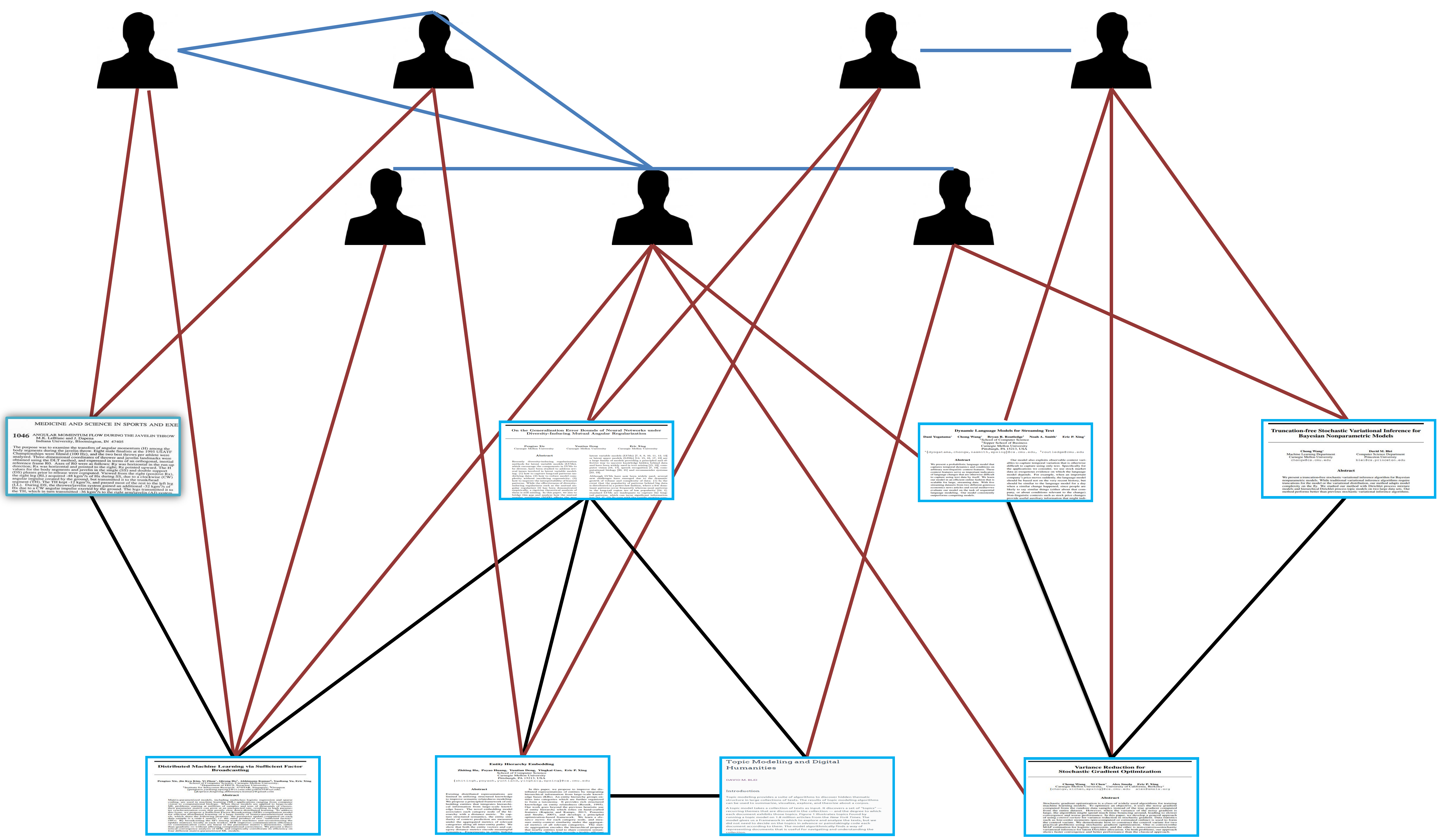}
\caption{The HMRN in two layers}
\label{fig:HINexample2}
\end{subfigure}
\caption{An example of HMRN: An author-paper network. (a) Papers are usually written by more than one authors. Such heterogeneous links between authors and papers are explicitly added to the two types of homogenous links - one is co-authorship links, and the other is paper-relevant links. (b) The author-paper network is reorganized into two layers to show its homogeneity and heterogeneity clearly. Within each layer, it has homogeneous links, e.g., author-author relationship colored in blue and paper-paper link colored in black. Between layers, there are heterogeneous links, i.e., author-paper links colored in red.}
\label{fig:HINexamples}
\end{figure*}

        Most studies on community detection have been focusing on homogeneous networks, where only one type of objects and single-relational links are considered \cite{newman2004detecting}\cite{lancichinetti2009community}. However, HMRNets with more than one types of objects and multi-relational links are ubiquitous in real-world scenarios. 
Take a bibliographic network as an example, shown in Figure \ref{fig:HINexamples}.
It contains two types of objects, i.e., authors and papers, and three types of relationships, i.e., co-author relationships (or social friendships), paper citation relationships (or paper similarity relationships), and author-paper relationships. Similar examples can also be found in social networks \cite{CantadorRecSys2011,zeng2013recommending} and biological networks \cite{sun2012will,tanay2004revealing,reimand2008graphweb}.

Heterogeneous information in networks might provide useful clues for community detection. However, it has only been partially exploited \cite{liu2014framework,yang2018meta}.
One branch of works utilizes the heterogeneous information implicitly by integrated it into a homogeneous network and then applies mature techniques developed for purely homogeneous networks for community detection \cite{wu2013community,sun2009rankclus}.
Another branch focuses on explicitly exploiting heterogeneous information only in subnetworks of HMRNet.
Two types of subnetworks are mainly studied.
One is homogeneous multi-relational networks which have only one type of nodes but with multiple types of links among them.
For instance, Facebook users could be friends of each other. Direct friendship links are established via this relationship.
 They could share posts of each other, via which indirect common interest links are established.
Another type of subnetworks is heterogeneous single-relational networks (HSRNet), such as bipartite networks and generalized $k$-partite networks.
Recently, Liu \textit{et al.} \cite{liu2014framework} attempt to compose a general community detection framework for HMRNets based on the above work on subnetworks.
It decomposes an HMRNet into three types of subnetworks, namely, unipartite graphs, bipartite graphs and $k$-partite graphs,
and then combine them into a unified framework via a composite modularity optimization formula.
However, how to weigh the modularities for different subnetworks is a tricky and challenging problem.
To sum up, no existing approaches could fully exploited an HMRNet yet. 

        To fill this gap, in this paper we propose a framework that could fully exploits the heterogeneous information in an HMRNet for community detection.
Inspired by the above work, we view an HMRNet as two types of subnetworks, i.e., homogeneous single-relational subnetworks and heterogeneous single-relational subnetworks.
In a homogeneous single-relational subnetwork, we follow the line of affinity propagation (AP) method.
In a heterogeneous single-relational subnetwork, we design an information exchange mechanism to mutually affect and enhance the AP results.
We combine the above two types of subnetworks by a maximum a posteriori (MAP) formula via a factor graph.
There are two types of factor nodes in the graph.
Homogeneous factor variables link to nodes of objects of the same type, and are functioned as collecting local homogeneous evidence.
Heterogeneous factor variables link to nodes of objects of different types. They build an `information bridge' among homogeneous single-relational networks.
By this, the clustering tendencies in one homogeneous layer can be passed through to its connected counterparts, and thus to affect the results of its connected counterparts. Ultimately, all community detection results would be enhanced.
A message passing procedure is derived to find a solution of this formula.
Finally, the practical superiority of our proposed method is verified on both synthetic networks and real-world networks including DBLP and Delicious-2K datasets.

        The remainder of this paper is organized as follows.
Section \ref{sec:relatedWork} reviews related work.
The following two sections introduce the proposed model and its inference algorithm.
Section \ref{sec:experiment} reports experimental results on a synthetic network and two real-world networks including DBLP and Delicious-2K datasets. Finally,
Section \ref{sec:conclusion} concludes this paper and discusses future work.

\section{Related Work}
\label{sec:relatedWork}

        Community structures is the organization of nodes into clusters with dense edges within clusters and comparatively sparse edges between clusters.
        It is one of the most relevant features of various types of information networks, such as social networks \cite{du2007community,pizzuti2008ga,nguyen2014dynamic,zhang2013predicting,kong2013inferring,zhang2014transferring,zhan2015influence}, the Internet and the World Wide Web networks \cite{boccaletti2006complex}, biological networks \cite{fortunato2010community}.
 Community detection is to identify such clustering structures. It is a fundamental step in disclosing rich semantic information in networks, such as finding circles of friends with common interests and discovering groups of researchers with similar research interests.

\noindent \textbf{Homogeneous networks}

        Most community detection work have been focusing on homogeneous networks \cite{lancichinetti2009community,leskovec2010empirical}. They can be roughly organized into four categories according to the criterion measuring how well a community structure is.
Modularity is a widely used criterion. It measures the strength of division of a network into communities by comparing the concentration of edges within communities with the random distribution of links \cite{Newman2004Physical,Mingming2014TransSS}.
A variety of approaches to obtain the maximal modularity have developed, such as simulated annealing \cite{good2010performance}, spectral optimization \cite{newman2013spectral}, the Louvain method \cite{blondel2008fast}, to name a few.
However, the resolution limit issue has been one notable drawback of this category \cite{lancichinetti2011limits}.

Similarity is another practical measurement for community detection. It quantifies pairwise closeness of nodes in networks and groups nodes that are similar to each other by maximizing overall similarities. Once the similarity is established, traditional clustering methods can be easily adapted. Methods in this category include hierarchical clustering \cite{newman2004detecting,murtagh2012algorithms}, spectral clustering \cite{van2013community,tsironis2013accurate}, AP \cite{frey2007clustering}. Note that AP is deemed superior considering it does not need a pre-defined number of communities.

Likelihood is a basic criterion for generative models. This category assumes a generative model for community structures and measures how likely an observed network is generated by it. The popular stochastic blockmodels (SBM) \cite{karrer2011stochastic,airoldi2009mixed} is under this category. Maximum likelihood estimation (MLE) and maximum a posterior (MAP) are two popular schemes related to this category and are used to learn parameters of the generative models.

Another popular criterion is betweenness-centrality. It is defined on each edge by counting the number of passing-by paths `between' pairs of nodes within networks \cite{girvan2002community}.
The communities within a network are identified by an iterative procedure. Within each step it alternate between removing edges with large `betweenness-centrality' and recomputing the measurement. Because recomputing is quite time-consuming, this category is generally impractical, even for a middle-sized network.

\noindent \textbf{Heterogeneous networks}

Earlier works exploit heterogeneous information along the lines developed for homogeneous networks \cite{shi2015survey}.
They focus on developing strategies to integrate heterogeneous qualities into homogeneous networks and then taking advantage of the sophisticated frameworks already existed for homogeneous networks.
For example, Tang \textit{et al.} developed four integration strategies to combine multi-relational structures into a unified single-relational network framework \cite{tang2012community}.
Despite the straightforwardness of this integrating scheme,
it usually cannot handle complex multi-relational links and may miss important clues for community detection.

Partial heterogeneous information has been exploited in the form of subnetworks, most of which are based on the prevalent modularity maximization method. 
Mucha \textit{et al.} developed a generalized modularity, called stability, for community detection in homogeneous multi-relational subnetworks.
Barber \cite{barber2007modularity} explored the extension of modularity maximization method for community detection in heterogeneous single-relational subnetworks.
Murata \cite{murata2010detecting} focused on developing a modularity maximization based co-clustering model for $k$-partite subnetworks.
As those methods utilize only partial of the heterogeneous information, useful clues for community detection might be ignored. 

Exploring the whole heterogeneous information has also been studied recently.
Liu \textit{et al.} \cite{liu2014framework} composed a general modularity optimization formula by decomposing HMRNet into subnetworks of three types, i.e., unipartite, bipartite and $k$-partite graphs. It considers all heterogenous information via distributing it into subnetworks.
However, how to weigh the modularities for subnetworks is still a tricky and challenging problem.
By contrast, MetaFac (MF) \cite{lin2009metafac,lin2011community} combines heterogeneous links into homogeneous ones via a tensor representation and treats these two different types of information equally.
Another example is in \cite{comar2012framework}, heterogeneous links are explored under a nonnegative matrix factorization (NMF) framework to discover implicit correspondence between subgroups of homogeneous networks.

The proposed framework is different with all above mentioned methods in the way of exploring heterogeneous information.
It explores these information in a more direct way.
Instead of trying to unify homogeneous information and heterogeneous information, our proposed method exploits them via explicit different strategies.
Homogeneous information is used for community detection for single homogeneous network, while heterogeneous information is used for information exchange between connected homogeneous networks. Via this, no weights need to be manually adjusted. In addition, the information exchange scheme would benefit the community detection for all homogeneous networks by mutually influence each other.

\begin{figure*}
\centering
\includegraphics[width=0.85\textwidth]{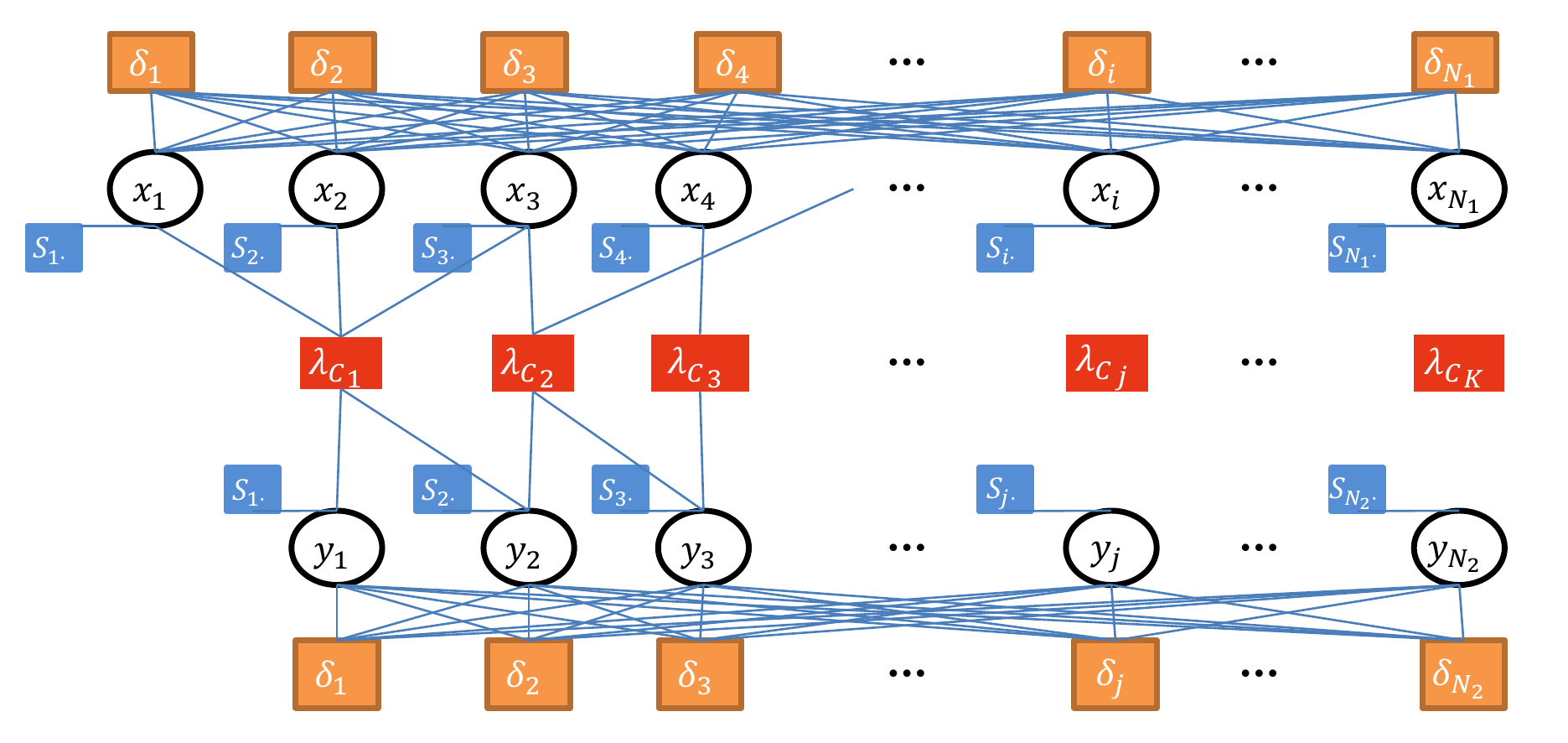}
\caption{The Factor Graph for Community Detection in HMRNet.}
\label{fig:factorgraphforwhole}
\end{figure*}

\section{A Factor Graph for Community Detection in HMRNet}
\label{sec:model}

        First, we set up the problem of community detection in HMRNet.
        Suppose an HMRNet $G=\langle V,E \rangle$.
        Without loss of generality, we focus on HMRNets with two types of objects and three types of links in this study.
        It can be easily extended to more types of objects and links.
 Two types of objects are $G(V)=V_{X} \cup V_Y$, and three types of links are $G(E) = E_{XX} \cup E_{XY} \cup E_{YY}$.
 These links connect both homogeneous and heterogeneous objects, i.e., $E_{XX} = \{\langle v_{i},v_{j} \rangle|v_{i},v_{j} \in V_X\}, ~E_{YY} = \{\langle v_{i},v_{j}\rangle|v_{i},v_{j} \in V_Y\},~E_{XY} = \{\langle v_{i},v_{j}\rangle|v_{i}\in V_X, v_{j} \in V_Y\}$.
Given HMRNet $G$, detecting communities is to divide objects of the same type into disjoint groups, where objects in the same group have similar properties.


        Take the above bibliographic network as an example again.
This HMRNet is equivalently re-organized into a two-layered representation as shown in Figure \ref{fig:HINexample2}.
The HMRNet is built with two homogeneous networks of two types' objects and heterogenous links in between.
The task of community detection is to simultaneously partition the author layer into communities with different research interests and the paper layer into groups with different research topics.

\noindent \textbf{AP for HSRNets}

        Our method follows the line of AP. It is prevalent for exemplar-centred clustering in both science and engineering area.
Considering the fact that detecting communities in networks is actually a task of clustering over network nodes, AP has been extended to network data. But most of them handle with homogeneous single-relational networks (HSRNets) only \cite{liu2008community,Hongwei2011IJCCSO}.
In these works, the objective of AP is to identify disjoint groups with centred exemplars, where the members in each group have the highest similarities to its own exemplar than to exemplars of other groups. As similarity is a core concept, it is usually measured by the topological structure of a network.
In this paper, we take the shortest distances between pairs of nodes as the similarity measurement.

Suppose $X=\{x_1,...,x_{N_1}\}$ be hidden variables representing possible exemplars in an HSRNet, and use $s(i,x_i)$ to serve as the pairwise similarity between object $v_i$ and its exemplar $x_i$. The objective of AP is then formulated as maximizing the overall pairwise similarities through the whole network. Formally,

\begingroup
\allowdisplaybreaks
\begin{align}
\max_X \mathcal{S}(X) &= \max_X
-\mathcal{E}(X)+ \Delta_{N_1}(X)
\label{eq:objectivefunction1},
\\
\mathcal{E}(X) &= -\sum_{i=1}^{N_1} s(i,x_i),  \notag 
\\
\Delta_{N_1}(X)&=\sum_{i=1}^{N_1} \delta_i(X), \notag \\
\delta_i(X) &=
\begin{cases}
-\infty, \mathrm{if} ~ x_i \neq i, \mathrm{but} ~\exists j: x_j = i;\\
0, \quad~ \mathrm{otherwise.}
\end{cases} \notag
\end{align}%
\endgroup
Here, $\mathcal{S}(X)$ represents the overall similarity to be maximized.
$\{\delta_i\}_{i=1}^{N_1}$ are delta functions used as constraints to guarantee that an object chosen by other objects as exemplar should be exemplar of itself first.

        The above objective function can be represented by a factor graph equally, as shown in the upper part of Figure \ref{fig:factorgraphforwhole}.
In this factor graph,
hollow circles represent exemplar variables, labelled by $x_i$.
Blue squares labelled by $S_{i\cdot}$ represent factor variables encoding local evidence.
Note that this evidence involves with only one variable $x_i$, and calculates the similarity between object $v_i$ and its potential exemplar $x_i$. As mentioned above, this similarity is associated with the shortest distance between them.
Orange squares labelled by $\delta_i$ represent factor variables encoding self-exemplar constraints. As indicated by dense connections between variable nodes and factor nodes, these constraints involve with all variables.
Equivalently, the objective function in Eq. \eqref{eq:objectivefunction1} can be obtained by summarizing over all factor nodes.
One optimal solution of this maximization problem is usually inferred via MAP, which is approximately achieved by a max-sum message passing procedure \cite{yedidia2000generalized}.

\noindent \textbf{Our method}

Based on AP for HSRNet, community detection in HMRNet is to discover exemplar-centred communities for its all homogeneous single-relational constituents. One HSRNet with a two-layered representation is shown in Figure \ref{fig:HINexample2}.
In addition to the homogeneous single-relational networks,
heterogeneous single-relational links might also carry significant information to improve the community detection results.
Intuitively, the papers, co-authored by researchers belonging to one community, should be grouped together with a high probability.
To utilize this information from heterogeneous links, we introduce a mediate information exchange mechanism. We detail it next, and derive an AP-like message passing procedure to infer an optimal community configuration in HMRNet in next section.

How to extract information from heterogeneous single-relational links to facilitate community detection in each homogeneous network is the key modeling problem to be addressed.
We apply the concept biclique in the bipartite graph theory \cite{bein2008clustering} to do this.
Specifically, in a two-layered HMRNet $G(V,E)=\langle V_X \cup V_Y, E_{XX}\cup E_{XY} \cup E_{YY}\rangle$,
a biclique $C = \{V_X^c \cup V_Y^c \}$ with two
subsets $V_X^c\subseteq V_X$ and $V_Y^c\subseteq V_Y$ should satisfy the following condition,
\begin{align*}
\forall v_i \in V_X^c,\forall v_j \in V_Y^c, ~\mathrm{then}~ \langle v_i,v_j \rangle\in E(V_X,V_Y).
\end{align*}
It is easily seen that every node in the first set $V_X^c$ is required to connect to every node in the second set $V_Y^c$.
In other words, a blockwise relationship between two layers is established.

To utilize this relationship to facilitate the community detection in HMRNet,
we assume that if all objects in set $V_X^c$ are grouped in one community, then all nodes in set $V_Y^c$ should also belong to the same community with a high probability.
Based on this assumption, the community structure information is exchanged between homogeneous layers via heterogeneous links.
Because the biclique assumption is quite strong and local, it is believed to benefit the detection on both homogeneous sides.

\begin{figure*}[tbh]
\centering
\begin{subfigure}{0.23\textwidth}
\includegraphics[width=\textwidth]{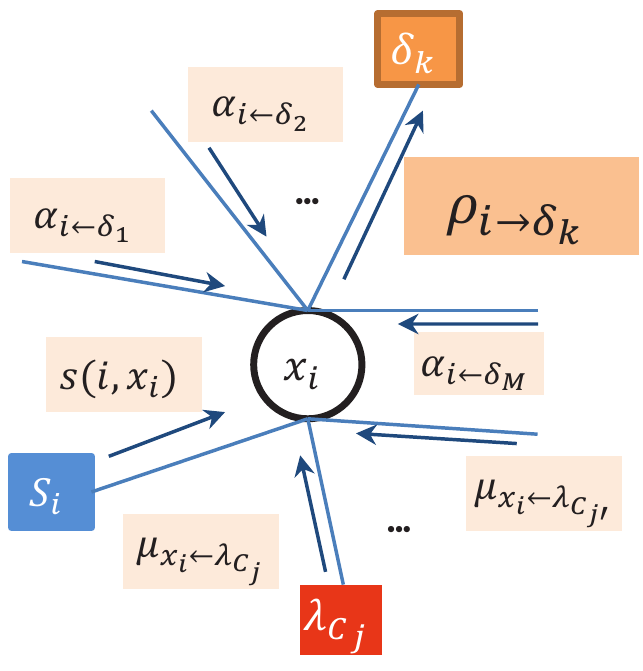}
\caption{Responsibility}
\label{fig:variablefactor1}
\end{subfigure}
\hspace{40pt}
\begin{subfigure}{0.24\textwidth}
\includegraphics[width=\textwidth]{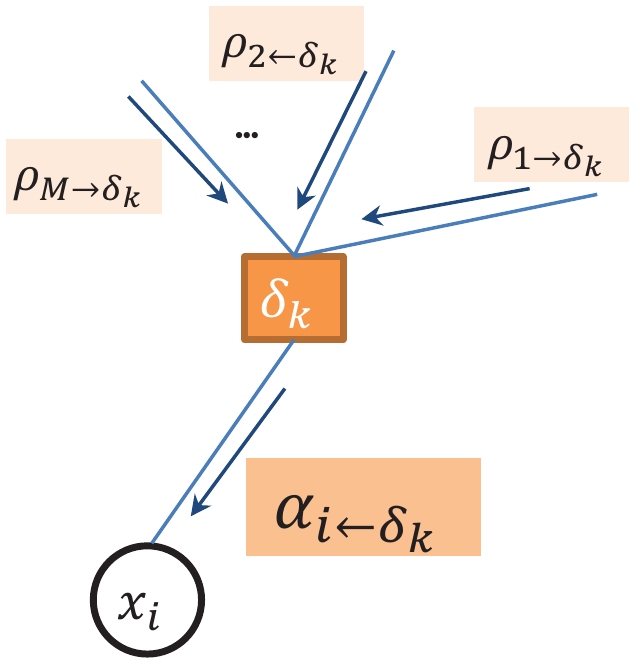}
\caption{Availability}
\label{fig:factornode1}
\end{subfigure}\\
\begin{subfigure}{0.24\textwidth}
\includegraphics[width=\textwidth]{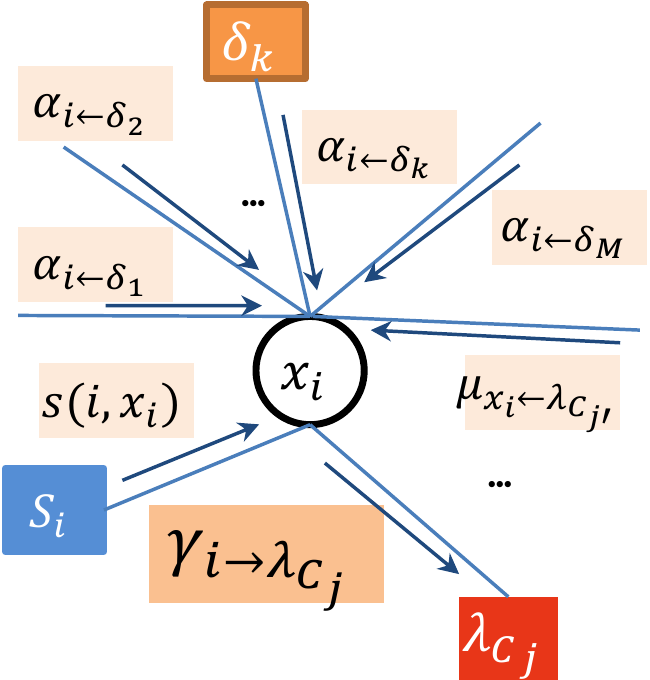}
\caption{Hub-collecting}
\label{fig:variablefactor2}
\end{subfigure}
\hspace{40pt}
\begin{subfigure}{0.24\textwidth}
\includegraphics[width=\textwidth]{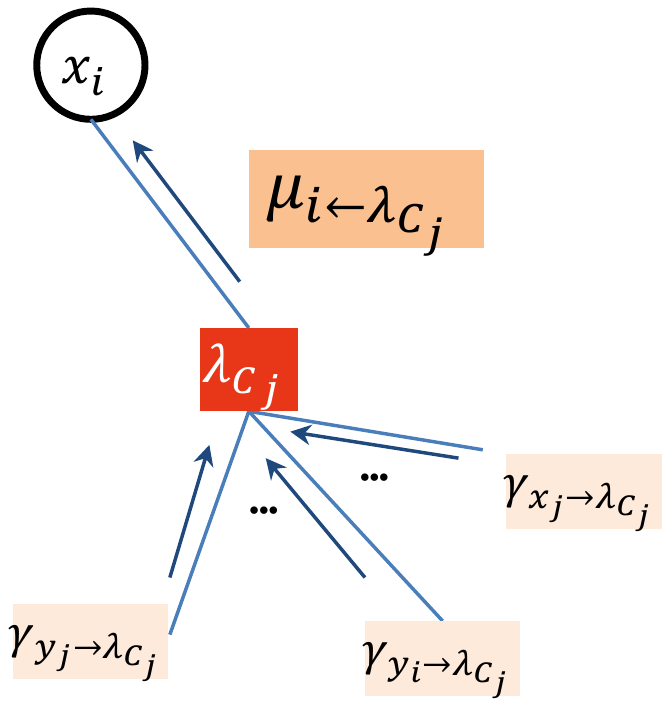}
\caption{Hub-broadcasting}
\label{fig:factornode2}
\end{subfigure}
\hspace{40pt}
\begin{subfigure}{0.24\textwidth}
\includegraphics[width=\textwidth]{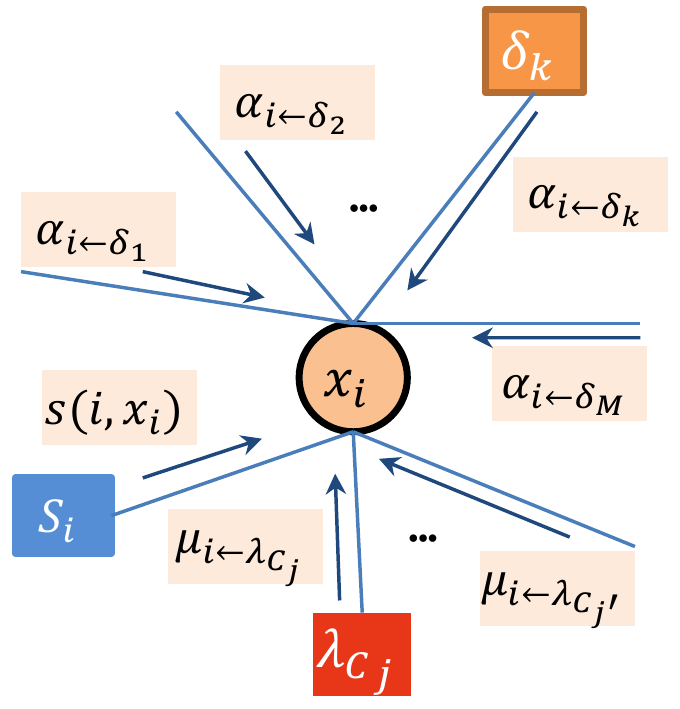}
\caption{MAP}
\label{fig:marginalEstimation}
\end{subfigure}
\caption{Messages in the Message Passing Algorithm}
\label{fig:messagePassing}
\end{figure*}

Combining the extra information from heterogeneous links into the homogeneous similarity measurements,
the new objective function for community detection in a two-layered HMRNet is formulated as
\begingroup
\allowdisplaybreaks
\begin{align}
& \mathcal{S}(X,Y) \notag \\
 &=
-\mathcal{E}(X)+ \Delta_{N_1}(X)
-\mathcal{E}(Y)+ \Delta_{N_2}(Y)
+ \Lambda_K(C) \label{eq:objectivefunction}, \\
& \Lambda_K(C) = \sum_{j=1}^K \lambda(C_j), \notag \\
& \lambda(C_j) =
\begin{cases}
0, \mathrm{if}~ x_i =k, \forall x_i \in C_j \mathrm{and}~ y_i=k', \forall y_i \in C_j; \\
-M, \quad \mathrm{otherwise}.
\end{cases}
\notag 
\end{align}%
\endgroup
$\mathcal{E}$ and $\Delta$ are homogeneous similarity functions and self-exemplar constraints respectively, the same as above.
$\lambda(C_j)$ is an information exchange function over a biclique $C_j$.
From above definition, when the objects in each side of a biclique are grouped together, namely grouped into the same community,
no penalty will apply.
Otherwise, a penalty measured by $M$ will take effect.
Note that $M$ is the only parameter added to the original AP model. It is a trade-off between the homogeneous similarity measure and the heterogeneous biclique information in obtaining an optimal community configuration.
Bicliques are basic structures to our model.
To find them, we applied an existing algorithm in \cite{alexe2004consensus}.

        The factor graph corresponding to the above formula is shown in Figure \ref{fig:factorgraphforwhole}.
The setting for homogeneous layers in HMRNet with similarity factors $S_{i\cdot}$ and self-exemplar constraint factors $\delta_i$ is the same with the original AP model.
The biclique-related factor $\lambda_{C_i}$ - encoding the heterogeneous links in biclique $C_i$ - is used to mutually enhance the community detection in the associated homogeneous networks.

\section{A Message Passing based Algorithm}
\label{sec:algorithm}

A message passing based algorithm inferring an optimal community configuration in HMRNet is presented in this section. The derived algorithm is symbolized with ``MP'' for short.
It is derived from a max-sum line \cite{koller2009probabilistic} to solve the maximization objective in (\ref{eq:objectivefunction}) by alternately propagating messages from variable nodes to factor nodes and vice versa.
The messages propagate within homogenous layers and between one homogeneous layer and its neighbouring heterogeneous factors $\Lambda_{K}$.
There are in total four types of messages with different functionalities with names `Responsibility', `Availability', `Hub-Collecting', and `Hub-Broadcasting'.
They are illustrated in Figure \ref{fig:variablefactor1} to Figure \ref{fig:factornode2} respectively.
Below we present the message formulas one by one, and leave their derivation to the Supplement.
Since there is no substantial differences between messages' formulas for different homogeneous layers,
we simply choose the layer $X$ shown in Figure \ref{fig:factorgraphforwhole} for demonstration.


\noindent\textbf{Responsibility messages}
\begingroup
\allowdisplaybreaks
\begin{align}\label{eq:rhoUpdate}
&\tilde{\rho}_{i \rightarrow \delta_k}(x_i = k) = s(i,x_i = k) \nonumber\\
& + \sum_{j:x_i \in C_j}\tilde{\mu}_{i \leftarrow \lambda_{C_j}}(x_i = k)
- \max\limits_{x_i = k', x_i \neq k} \left\{ s(i,x_i = k')\right. \nonumber\\
& + \left. \tilde{\alpha}_{i \leftarrow \delta_{k'}}(x_i = k')
+ \sum_{j:x_i \in C_j} \tilde{\mu}_{i \leftarrow \lambda_{C_j}}(x_i = k') \right\}.
\end{align}
\endgroup
A `responsibility' message, from variable node $x_i$ to self-exemplar constraint factor node $\delta_k$,
reflects the evidence of object $i$ choosing object $k$ as its exemplar, i.e., $x_i=k$ by competing with its biggest competitor.
The competitive power is based on the sum of the similarity between object $i$ and the competitor itself $k',k'\neq k$, the availability of the competitor, and the support from competitor-related heterogeneous bicliques.

\noindent\textbf{Availability messages}
\begingroup
\allowdisplaybreaks
\begin{equation}
\begin{aligned}
&\tilde{\alpha}_{i \leftarrow \delta_k} (x_i=k) = \\
&
\begin{cases}
\sum_{i':i'\neq i} \max\{0,\tilde{\rho}_{{i'}\rightarrow \delta_k}(x_{i'}=k)\},
\quad \quad\quad \quad \mathrm{for}~k = i; \\
\min\left\{ 0,
\tilde{\rho}_{k\rightarrow \delta_k}(k) + \sum\limits_{i'\neq \{i,k\}}~\max\{0,\tilde{\rho}_{{i'}\rightarrow \delta_k}(x_{i'} = k)\}\right\},
\\
\quad\quad \quad\quad\quad\quad\quad\quad\quad\quad\quad\quad\quad\quad\quad\quad\quad\quad
\mathrm{for}~k \neq i.
\end{cases}
\end{aligned}
\label{eq:alphaUpdate}
\end{equation}
\endgroup
An `availability' message, from self-exemplar constraint factor node $\delta_k$ to variable node $x_i$,
shows the opportunity of object $k$ chosen by object $i$ as its exemplar, by taking into account the accumulating scores of responsibilities of the exemplar $k$ for all the other objects.

\begin{figure*}[tbh]
\hspace{-10pt}\begin{subfigure}{0.35\textwidth}
\includegraphics[width=\textwidth]{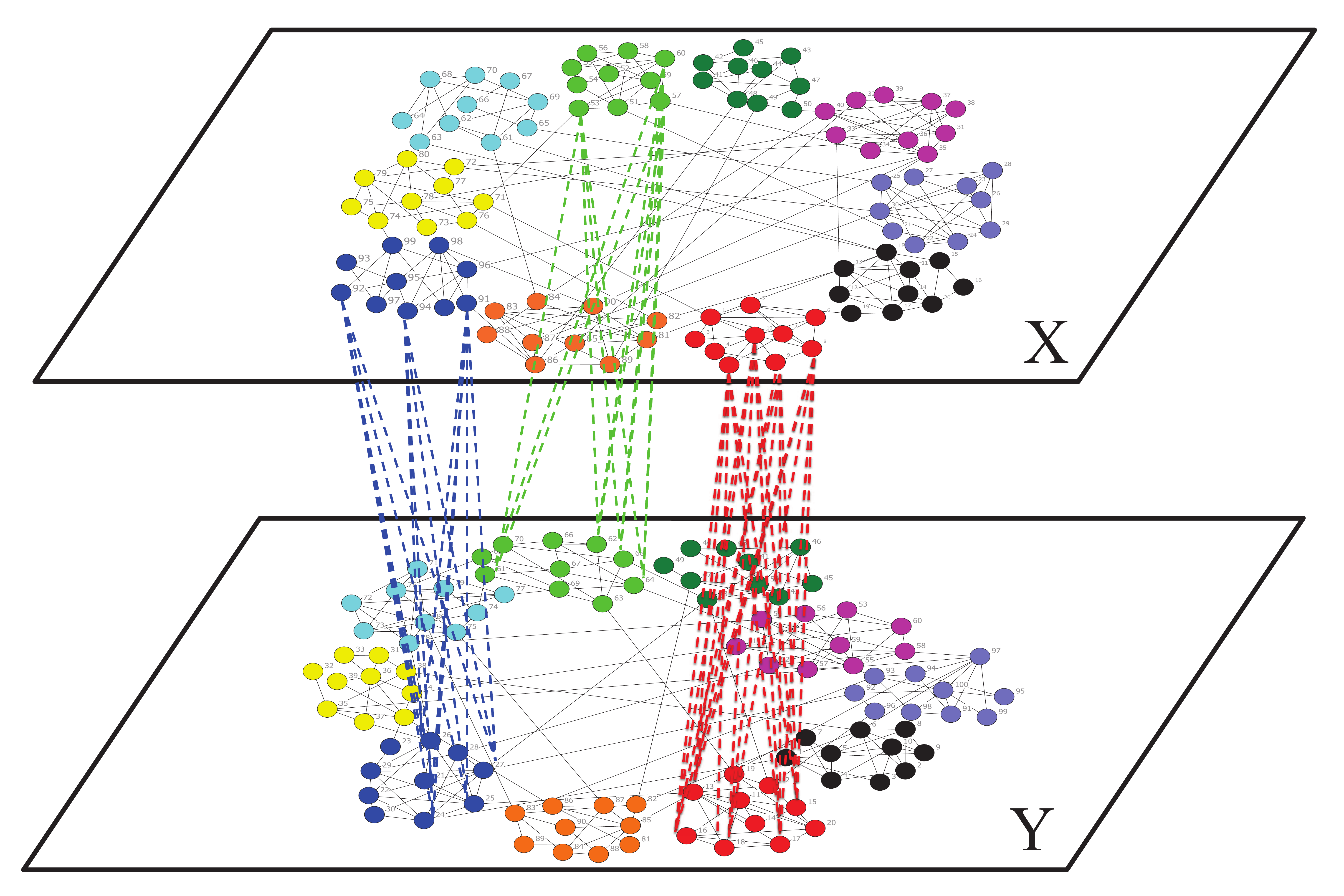}
\caption{Ground-Truth}
\label{fig:syntheticGroundTruth}
\end{subfigure}
\hspace{-10pt}
\begin{subfigure}{0.35\textwidth}
\includegraphics[width=\textwidth]{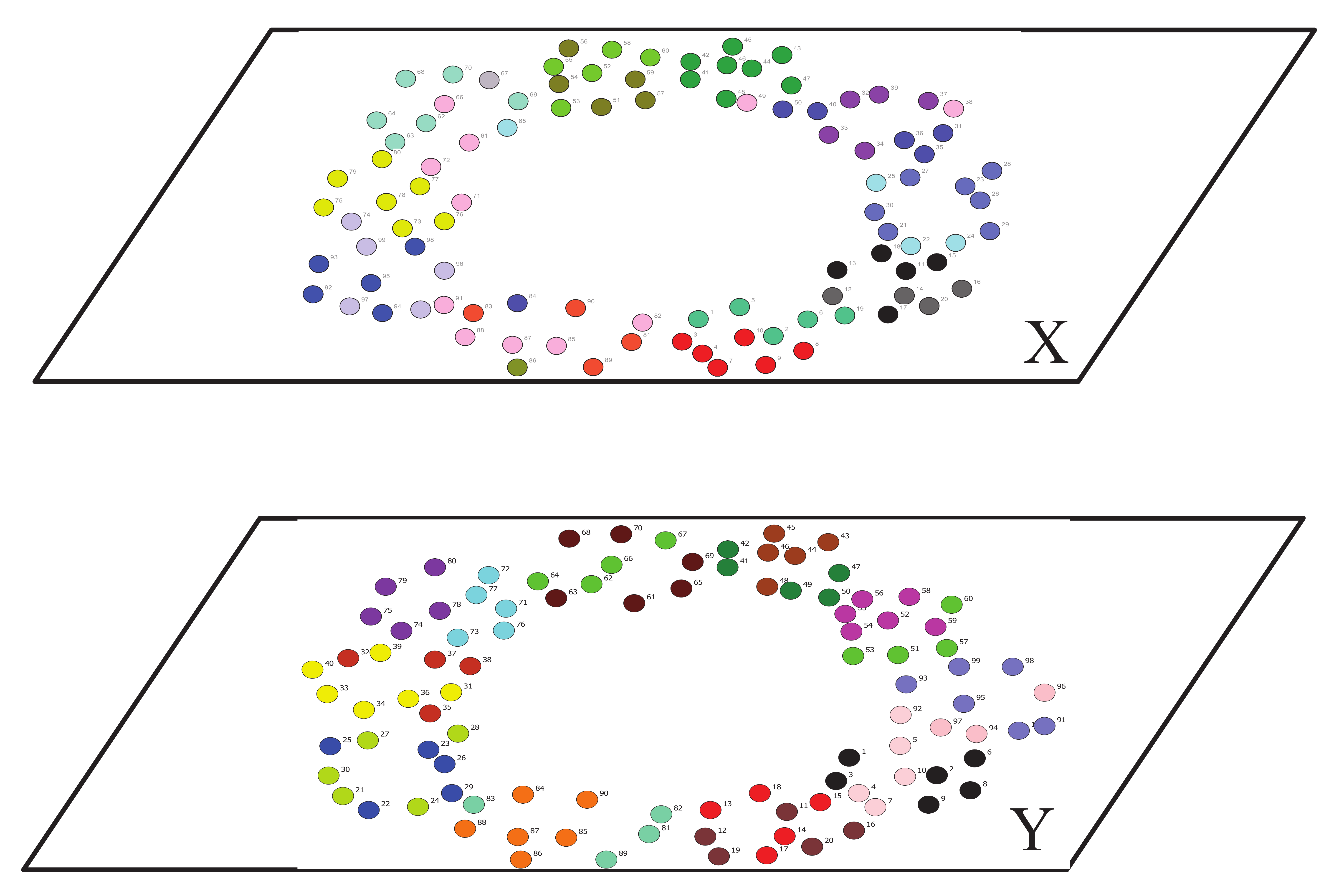}
\caption{AP Results}
\label{fig:syntheticAPY}
\end{subfigure}
\hspace{-10pt}
\begin{subfigure}{0.35\textwidth}
\includegraphics[width=\textwidth]{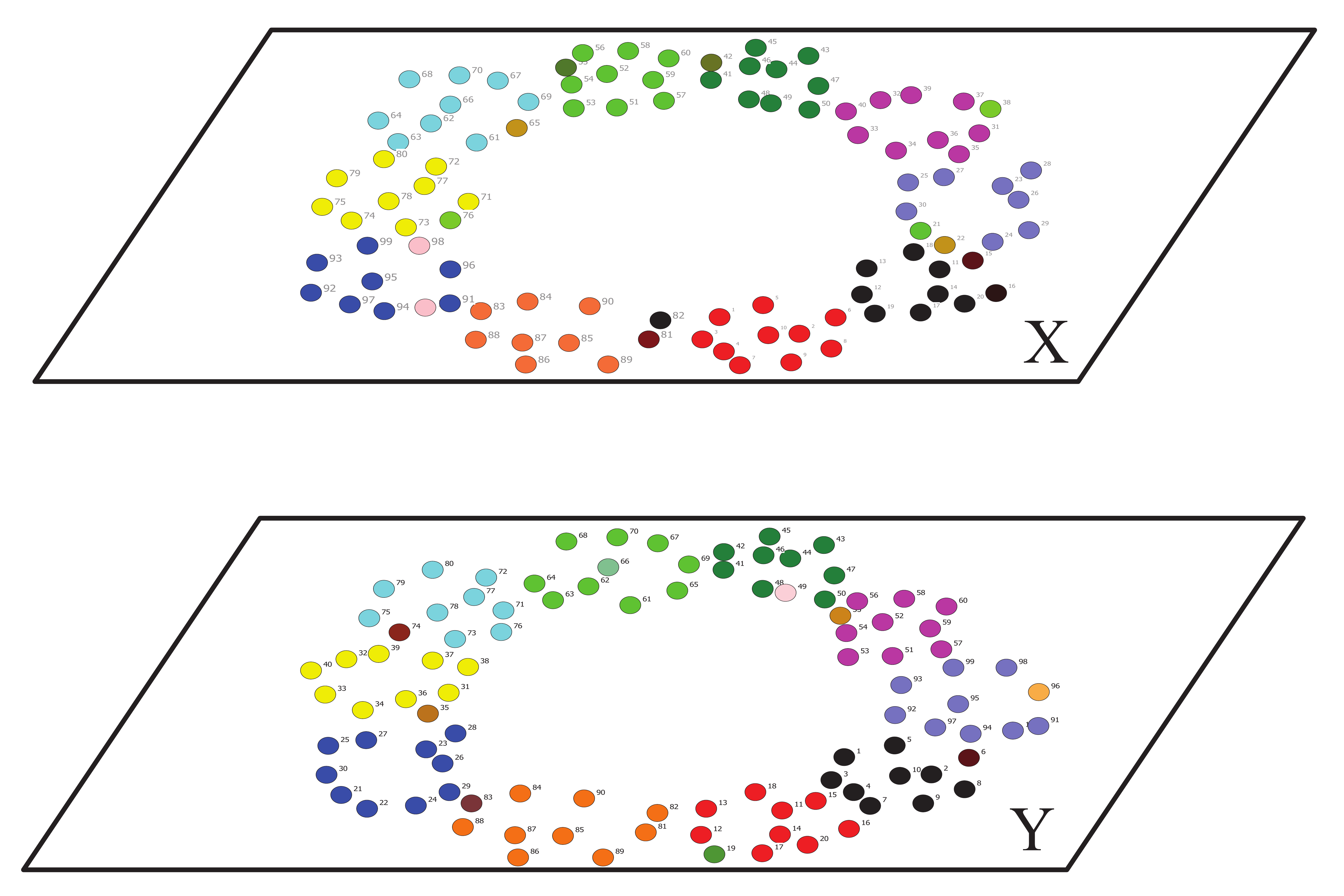}
\caption{MP Results}
\label{fig:syntheticBPY}
\end{subfigure}
\caption{A Qualitative Example of Experimental Results for Community Detection in HMRNet}
\label{fig:syntheticFigR}
\end{figure*}

\noindent\textbf{Hub-collecting messages}
\begin{align}
\tilde{\gamma}_{i \rightarrow \lambda_{C_j}}(x_i = k') &= s(i,x_i = k') + \tilde{\alpha}_{i \leftarrow \delta_{k'}}(x_i=k') \nonumber \\
& + \sum_{\substack{j':j'\neq j,\\ x_i\in C_{j'}}} \tilde{\mu}_{i \leftarrow \lambda_{C_{j'}}}(x_i).
\label{eq:gammaUpdate}
\end{align}
A `hub-collecting' message is sent from a variable node $x_i$ to a heterogeneous biclique factor node $\lambda_{C_j}$.
As its name indicates,
a biclique factor node $\lambda_{C_j}$ functions as a ``hub'' and collects
messages from all variable nodes belonging to $C_j$.
Each collected message comprises the similarity between the object $i$ and its chosen exemplar $k'$, the availabilities of exemplar $k'$ for the object $i$, and the broadcasting messages from all other heterogeneous biclique factor nodes that related to objects $i$.

\noindent\textbf{Hub-broadcasting messages}
\begingroup
\allowdisplaybreaks
\begin{align}\label{eq:muUpdate}
&\tilde{\mu}_{i \leftarrow \lambda_{C_j}}(x_i = k) = \nonumber\\
&
\begin{cases}
\sum\limits_{{i'}\in C_j^x} \max_{x_{i'}} \tilde{\gamma}_{i' \rightarrow \lambda_{C_j}} (x_{i'})+
\sum\limits_{{j'}\in C_j^y} \max_{y_{j'}} \tilde{\gamma}_{j' \rightarrow \lambda_{C_j}} (y_{j'})
, \\
\quad \mathrm{if}~x_{i'}^{\star} = k, ~\forall i' \neq i, i' \in C_j^x, x_{i'}^{\star} = \max_{x_{i'} }\tilde{\gamma}_{i' \rightarrow \lambda_{C_j}} (x_{i'}) \\
\quad \quad \mathrm{and} \quad y_{j'}^{\star} = k', ~\forall j' \in C_j^y,
y_{j'}^{\star} = \max_{y_{j'} }\tilde{\gamma}_{j' \rightarrow \lambda_{C_j}} (y_{j'}); \\
\quad
\\
\max \begin{bmatrix}
\sum\limits_{{i'}\in C_j^x} \max_{x_{i'}} \tilde{\gamma}_{i' \rightarrow \lambda_{C_j}} (x_{i'}) - M, \\
\sum\limits_{i' \in C_j^x} \tilde{\gamma}_{i' \rightarrow \lambda_{C_j}}(x_{i'} = k)
\end{bmatrix}, \\
\quad \mathrm{otherwise.}
\end{cases}
\end{align}
\endgroup
Here $C_j^x = C_j \backslash \{y_{\cdot}\}$, $\{y_{\cdot}\}$ is the subset from $Y$-layer and belongs to the biclique $C_j$.
A `hub-broadcasting' message is sent from a heterogeneous biclique factor node $\lambda_{C_j}$ to a variable node $x_i$.
After integrating all messages from its members the biclique factor node, functioned as a ``hub'', broadcasts its decision to variable $x_i$ that whether the community configuration $x_i=k$ needs to be penalised or not.

\noindent\textbf{MAP summarization}

Variable node $x_i$ takes all its into-messages into account
to decide its exemplar with the formula below, as shown in Figure \ref{fig:marginalEstimation}.

\begingroup
\allowdisplaybreaks
\begin{subequations}
\begin{align}
&\hat{x}_i= \arg\max\limits_{x_i=k} \nonumber \\
& \left\{s(i,x_i=k)
+ \sum_{j:x_i\in C_j}\tilde{\mu}_{i \leftarrow \lambda_{C_j}} (x_i=k) + \tilde{\alpha}_{i \leftarrow \delta_k}(x_i=k) \right\}
 \notag
\\
&= \arg \max_{x_i=k} \left\{ \tilde{\rho}_{i \rightarrow \delta_k}(x_i = k) + \tilde{\alpha}_{i \leftarrow \delta_k}(x_i=k) \right\}. \label{eq:thePesudoMar2}
\end{align}
\label{eq:thePesudoMar}
\end{subequations}
As Eq. \eqref{eq:thePesudoMar2} shows, the MAP objective chooses the exemplar for object $i$ with the highest sum of responsibility and availability.

Our algorithm is summarized in Algorithm \ref{alg:APforHIN} and proof details can be found in Supplement.

\begin{algorithm}[h]
\caption{Message Passing Algorithm for Community Detection in HMRNet}
\begin{algorithmic}[1]
\REQUIRE{Similarity matrices $S_X$ and $S_Y$, binary matrix of between-layer links: $E_{XY}$}, stop criterion parameter $n$, iteration index $\mathrm{iter} = 1$.
\STATE Initialize message matrix: $\tilde \rho_X$, 
$\tilde \rho_Y$, 
$\tilde \alpha_X$, 
$\tilde \alpha_Y$, 
$\tilde \gamma_X$, 
$\tilde \gamma_Y$, 
$\tilde \mu_X$, 
$\tilde \mu_Y$.
\STATE Find all bicliques $\{C_j,j={1:K}\}$ within $E_{XY}$.
\REPEAT
\STATE Update responsibility messages $\tilde \rho_X$ and $\tilde \rho_Y$ with Eq. \eqref{eq:rhoUpdate} along with $\{C_j\}$.
\STATE Update availability messages $\tilde \alpha_X$ and $\tilde \alpha_Y$ with Eq. \eqref{eq:alphaUpdate}.
\STATE Update hub-collection messages $\tilde \gamma_X$ and $\tilde \gamma_Y$ with Eq. \eqref{eq:gammaUpdate}.
\STATE Update hub-broadcast messages $\tilde \mu_X$ and $\tilde \mu_Y$ with Eq. \eqref{eq:muUpdate} along with the bicliques $\{C_j\}$.
\STATE
$\tilde P_X = \{ \tilde{\rho}_{i \rightarrow \delta_k}(x_i = k) + \tilde{\alpha}_{i \leftarrow \delta_k}(x_i=k)\}_{\substack{i=1:N_1\\k=1:N_1}}$;
~~\\
$\tilde{P}_Y = \{\tilde{\rho}_{i \rightarrow \delta_k}(y_i = k) + \tilde{\alpha}_{i \leftarrow \delta_k}(y_i=k)\}_{\substack{i=1:N_1\\k=1:N_2}}$;
\IF {isempty($\tilde P_X>0$) $\mid$ isempty($\tilde P_Y>0$)}
\STATE continue;
\ENDIF
\STATE
$[\sim,L_X(\mathrm{iter})] = \max(\tilde P_X,[],2)$;\\
$[\sim,L_Y(\mathrm{iter})] = \max(\tilde P_Y,[],2)$;
\STATE $\mathrm{iter} = \mathrm{iter} + 1$;
\UNTIL 
{$\operatorname{length}(\operatorname{unique}(L_X(\mathrm{iter}-n:\mathrm{iter}),\text{`rows'}))==1$}\\
\& $\operatorname{length}(\operatorname{unique}(L_Y(\mathrm{iter}-n:\mathrm{iter}),\text{`rows'}))==1$.
\RETURN Community configurations $L_X, L_Y$.
\end{algorithmic}
\label{alg:APforHIN}
\end{algorithm}

\noindent\textbf{Complexity analysis}

The main memory and computation consumption are from the four types of messages (Eq.\ref{eq:rhoUpdate},\ref{eq:alphaUpdate},\ref{eq:gammaUpdate},\ref{eq:muUpdate}) and the MAP summarization (Eq.\ref{eq:thePesudoMar}).
For the memory complexity, within an $l$th homogeneous layer $X_{l}$, each message assignment, i.e., $x_i=k$, requires all the other possible assignments of this message, i.e., $x_i = k'$ with $\{k'\neq k\}$ whose size is $N_{X_{l}}-1$ with $N_{X_l}$ the size of this homogeneous network. In other words, the memory complexity of each node is $\mathcal{O}(N_{X_{l}})$. Summarizing over all messages results in memory complexity of $\mathcal{O}(\sum_{l} N_{X_{l}}^2)$.

For computation complexity, the most time-consuming elementary calculation of our algorithm is $\mathrm{max}$ over all possible message assignments for each node as in \eqref{eq:rhoUpdate} and \eqref{eq:thePesudoMar2}.
Updating $\tilde \rho$ \eqref{eq:rhoUpdate} itself has the time complexity of $\mathcal{O}(N_{X_{l}})$ as there are $N_{X_{l}}$ possible message assignments. Besides, it results in just one item of the list of all possible message assignments for updating MAP summarization in \eqref{eq:thePesudoMar2}. As a result, updating MAP summarization for each node within each iteration takes the time complexity of $\mathcal{O}(N_{X_{l}}^2)$.
Finally, taking all nodes of the network and supposing $T$ iterations are required for algorithm convergence, our algorithm has the time complexity of $\mathcal{O}(T \max_{l} (N_{X_{l}}^3))$.

Note that within each of the $T$ iterations, our algorithm takes cubic time in terms of the size of a network, which is much higher than the linear form the baseline algorithm AP takes. We attribute this to exploring more information, i.e., heterogeneous bicliques, in our method, where the computation of each message assignment is not only dependent on its own-related information any more as in AP but also all the message assignments in the network. 

\section{experimental results}
\label{sec:experiment}

We evaluated the performance of the proposed method on both synthesized HMRNets with ground truth community structures and two real-world HMRNets without ground truth community structures, i.e., DBLP and Delicious-2K.

\noindent\textbf{Evaluation Metrics}

When ground-truth community structures known, three metrics are applied to evaluate clustering performance, i.e. accuracy, normalized mutual information (NMI), and variation of information (VI).

\noindent $\bullet$
Accuracy provides a straightforward way to measure how well a detected community structure matches the ground-truth.
Formally, given two aligned community structures (implemented by Hungarian algorithm), i.e., detected $X'$ and the ground-truth $X$, the accuracy is calculated by
\begin{align}
\mathrm{Accuracy}(X')=\frac{\sum_{i=1}^{|X|} \delta(x_i==x'_i) }{|X|} \times 100\%, \nonumber
\end{align}
with $x_i\in X$ and $x'_i \in X'$. $\delta(\cdot)$ is the Dirac delta function, and $|\cdot|$ calculates the cardinality of a set.

\noindent $\bullet$ The other two metrics - NMI and VI -  are based on information theory and can be explained as that NMI quantifies `similarity' between two community structures, while VI measures `dissimilarity' of them. Therefore, higher NMI values indicates better matching results, while VI exhibits an exactly opposite trend, namely the higher the VI values are,  the worse the matching results are. They
are respectively formulated as \cite{Mingming2014TransSS}
\begin{align}
\mathrm{NMI}(X,X')&=2\mathrm{I}(X,X')/(\mathrm{H}(X) + \mathrm{H}(X')) \nonumber
\end{align}
and
\begin{align}
\mathrm{VI}(X,X') &= \mathrm{H}(X) + \mathrm{H}(X')-2\mathrm{I}(X,X'), \nonumber
\end{align}
where $\mathrm{H}(\cdot)$ is the entropy function, and $\mathrm{I}(X,X') = \mathrm{H}(X) + \mathrm{H}(X') - \mathrm{H}(X,X')$ is the mutual information function. Both of them are formally calculated by
$
\mathrm{H}(X) = -\sum_{x_i \in X}\frac{|c_i|}{N}\log \frac{|x_i|}{N}$, and $
\mathrm{H}(X,X') =-\sum_{x_i \in X, x'_j \in X'} \frac{|x_i \cap x'_j|}{N} \log(\frac{|x_i \cap x'_j|}{N})$ with $N$ the cardinal number of $X$.

When no ground-truth is given, to measure the quality of a community structure, meta-data have been explored \cite{comar2012framework}. More commonly used information such as link density and separability of discovered communities are adapted here. Following \cite{yang2015defining}, four metrics are chosen here, i.e., modularity (Q), triangle participation ratio (TPR), conductance (C) and cut ratio (CR).

\noindent $\bullet$ Modularity (Q) \cite{lancichinetti2011limits,Newman2004Physicalfinding}, based on the justification that
a good community structure should have more intra-community links rather than inter-community ones, is formulated as \cite{Mingming2014TransSS}
\begin{equation}
Q(X) = \sum_{X_S \subset X} \left[ \frac{|E_{X_S}^{in}|}{|E|} - (\frac{2|E_{X_S}^{in}| + |E_{X_S}^{out}|}{2|E|})^2 \right], \nonumber
\end{equation}
with $X_S$ a community in $X$. $|E_{X_S}^{in}|$ is the number of links within community $X_S$, and $|E_{X_S}^{out}|$ counts links outgoing community $c_i$. $|E|$ is the number of links within the whole network.
When $Q$ approaching $1$, the community structure under examined shows strong community properties. On the contrary, when $Q$ approaching $0$, the given community structure is no better than a randomly generated network with no cluster structures. Typically,  $Q$ falls in $[0.3, 0.7]$ \cite{Newman2004Physicalfinding}.

\noindent $\bullet$ Conductance (C) measures the separability of a community via the fraction of outgoing link volume locally in the community, and is defined as
\begin{align}
\text{C}(X_S) = |E_{X_S}^{out}|/(2|E_{X_S}^{in}|+|E_{X_S}^{out}|). \nonumber
\end{align}
The smaller the overall conductance value of a community structure is, the better the local separability of communities is achieved.

\noindent $\bullet$ Triangle Participation Ratio (TPR) measures the density of a community via the fraction of triads within the community, and is formulated as
\begin{align}
\text{TPR}(X_S) =& |\{x_i \in X_S,\{  (x_j,x_k):x_j,x_k\in X_S, \nonumber \\
&(x_i,x_j),(x_j,x_k),(x_i,x_k)\in E_{X_S}  \}\neq \emptyset\}|  /{|X_S|}.
\nonumber
\end{align}
The bigger the overall TPR value of a community structure is, the denser the communities within it are.

\noindent $\bullet$ Similar to conductance, Cut Ratio (CR) also measures the separability of a community. However, it is achieved via a global view considering the whole community structure, which is attained via the fraction of observed links (out of all possible edges) leaving the community. Formally,
\begin{align}
\mathrm{CR}(X_S) =  |E_{X_S}^{out}| / (|X_S| \times (|X|-|X_S|)). \nonumber
\end{align}
The smaller the overall CR value of a community structure is, the better the global separability of communities is achieved.


\begin{table*}[tp]
\centering
\caption{\\Performance on the Synthetic Network I}
\renewcommand{\arraystretch}{1.3}
\begin{tabular}{r| lllll | lllll   } \hline
&\multicolumn{5}{c|}{network X} & \multicolumn{5}{c}{network Y} \\
& AP &MP\textsuperscript{*} & MF &EV &Truth &AP & MP\textsuperscript{*} &MF&EV& Truth \\ \hline
$\%$ &67.04 & \textbf{92.5} &91.54& 33.84& 100
& 62.54 &\textbf{81.0} &76.39&30.82& 100\\
\#  &17 &\textbf{10} &10 &10&10
& 18  &\textbf{12}&10&10 &10 \\
NMI &0.81 &\textbf{0.84}&0.83&0.16&1
 & 0.82 & \textbf{0.84} &0.83&0.21&1\\
VI & 0.97 & \textbf{0.75}&0.80& 2.74&0
 &0.94 &\textbf{0.74}&0.82& 2.63&0\\
$Q$ &0.54 & \textbf{0.64} &0.63&0.07& 0.68
 & 0.50 & \textbf{0.65} &0.65&0.03& 0.72\\
TPR\textsuperscript{\#} & 39.18 & \textbf{50.22} &49.78&0.08& 68.8
& 31.85 & 47.30  &\textbf{48.01}&0.07& 47.2  \\
$C$ & 0.75 & \textbf{0.73} &0.74& 0.85 &0.12
  & 0.83 & \textbf{0.71}&0.75& 0.89&  0.14 \\
CR & 0.35 & \textbf{0.35} &0.42&0.36&0.06
& 0.37 & \textbf{0.31}&0.44& 0.34& 0.06
\\ \hline
\multicolumn{7}{l}{\textsuperscript{*}\footnotesize{The results were achieved by setting $M$=1.3}, \textsuperscript{\#}\footnotesize{$\times 10^3$}. }\\
\end{tabular}
\label{tab:bestSynResults}
\end{table*}

\subsection{Synthetic Datasets}
\label{sec:synexperiment}

\begin{figure*}[ht]
\centering
\includegraphics[width=1\textwidth]{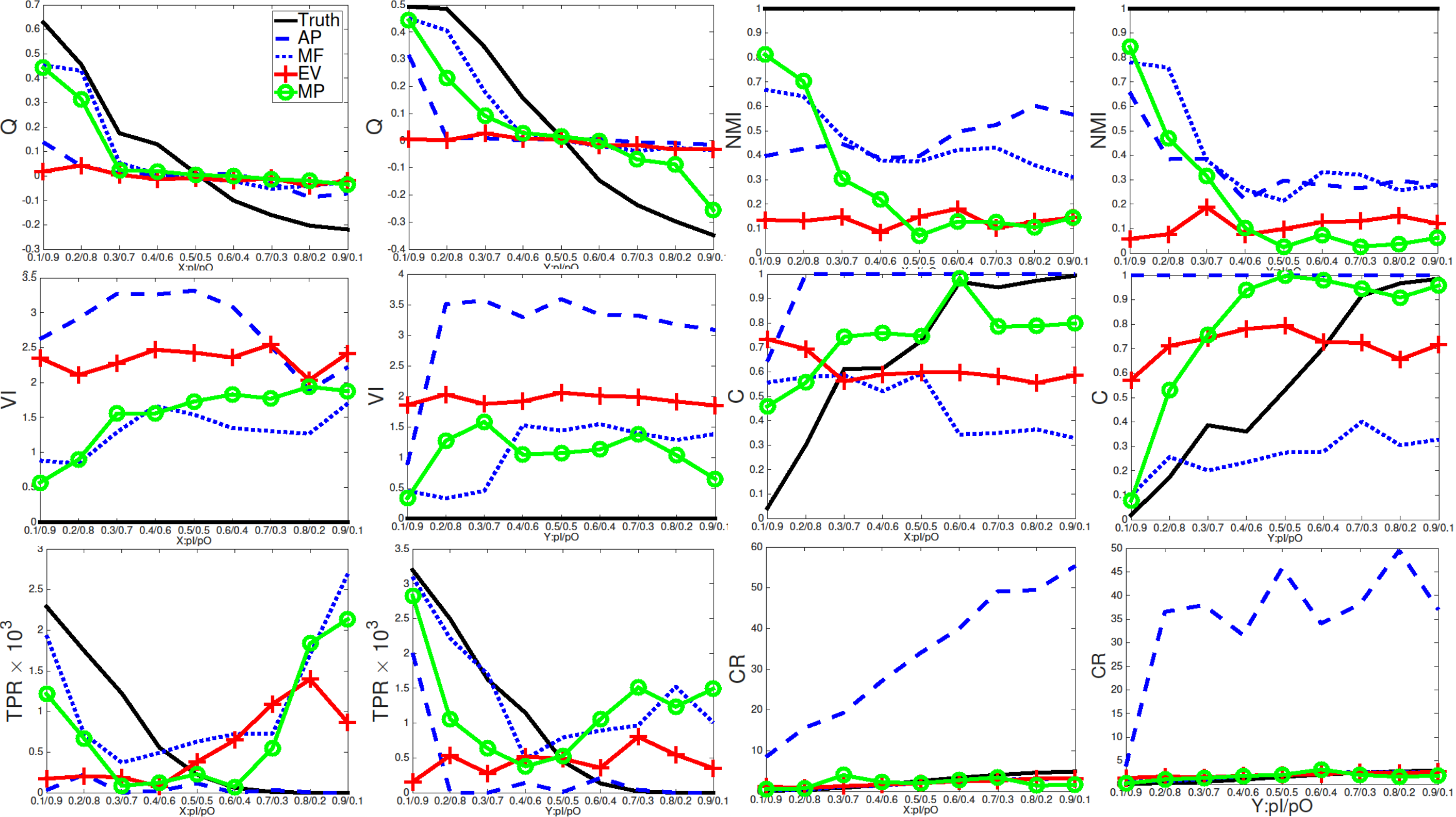}
\caption{{Performance on randomly generated synthetic networks with varying inter- and intra-community link generating probabilities, i.e., pI - the inter-community link generating probability and pO - the intra-community link generating probability. Six metrics are evaluated, i.e., modularity (Q), NMI, VI, conductance (C), TPR, and CR.}
}
\label{fig:synII}
\end{figure*}

\noindent\textbf{Synthetic network I}

A two-layered HMRNet with communities of equal size was generated by two steps.
First, two homogeneous networks were generated with fixed link generating probabilities. The probability for intra-community links is $0.85$ and the one for inter-community links is $0.15$. Each homogeneous layer contains $100$ nodes, and every $10$ nodes are clustered as a community.
Then, ten heterogeneous bicliques were generated randomly.
Two communities were randomly selected from each homogeneous network. Then two subsets from the two communities containing two to ten nodes were randomly chosen. Finally, these two subsets are fully connected to form a biclique.
A community structure generated from the above procedure is demonstrated in Figure \ref{fig:syntheticGroundTruth}. The nodes belonging to different communities were painted with different colours; the homogeneous single-relational edges were in grey; the edges belonging to fully connected bicliques were painted with the color of their two-end heterogeneous nodes (For clarity, only three bicliques are shown).

We study the effectiveness of the proposed algorithm both qualitatively and quantitatively by comparing it to the baseline algorithm AP.
Two community structures detected by the two algorithms are shown in Figure \ref{fig:syntheticAPY} and \ref{fig:syntheticBPY}.
As demonstrated, the community structure obtained by MP is far more close to the ground-truth than the one obtained by AP. We attribute this superiority to the heterogeneous information which are effectively explored by the proposed method but not by AP.


In addition to the baseline algorithm AP, we compare the proposed algorithm to two more state-of-the-art methods - MF and EV \cite{jia2010affinity}, which also dedicate to exploring heterogeneous information.
We emphasize that the number of communities is not required by AP and MP algorithms but needed by MF and EV, since the latter two cannot automatically determine this parameter. In the experiments, the performance was achieved with fixed $K=10$ for both MF and EV. The quantitative result is shown in Table \ref{tab:bestSynResults}.
Note that the trade-off parameter $M$ in MP is fixed to $1.3$, and the study to this parameter is introduced later in this section.
{It is easily seen that
the proposed method, i.e., MP, achieves consistently better results than the baseline method AP with a large margin in terms of the evaluation metrics and the closeness to the ground truth.}
The accuracy achieved by AP was improved by MP with $25.46\%$ for network $X$ and $18.46\%$ for network $Y$.
The numbers of communities for both networks found by MP were much closer to the ground-truth than AP.
Significant improvement of NMI and VI are also achieved by MP over AP.
According to modularity (Q), MP discovers communities with more cluster-like structures. From TPR, C, and CR, we claim that MP discovers community structure with denser intra-community links and better separable communities.


\begin{table*}[htp]
\centering
\caption{Performance on real-world dataset DBLP}
\renewcommand{\arraystretch}{1.3}
\begin{tabular}{r| llll | llll } \hline
&\multicolumn{4}{c|}{network Papers} & \multicolumn{4}{c}{network Authors} \\
&AP &MF&EV&MP\textsuperscript{*} &AP &MF&EV&MP\textsuperscript{*}  \\ \hline
$Q$ &0.6436 &0.6038&0.0714& \textbf{0.6460} &0.2980 &\textbf{0.3239} &0.2764&0.3010 \\
\#&4235 &2000&100&\textbf{2497} & 2002 &2000&100& \textbf{1981} \\
$C$ &0.8023&\textbf{0.7}&0.9248&0.7083 &0.8864 &0.8290&\textbf{0.2672}& {0.8292} \\
TPR\textsuperscript{\#}  & 1.42 &1.4340&0.0722&\textbf{1.66}&1.64 &\textbf{1.9055}&0.7214&{1.70} \\
CR &260.34&54.0624&\textbf{5.1377}&{56.43}&267.18 &256.4784&3.5783& \textbf{254.53}
\\ \hline
\multicolumn{9}{l}{\textsuperscript{*}\footnotesize{The results were achieved by setting $M$=0.1.
}
\textsuperscript{\#}\footnotesize{$\times 10^3$}. }
\end{tabular}
\label{tab:DBLP}
\end{table*}

\begin{table*}[htp]
\centering
\caption{Performance on real-world dataset Delicious-2K}
\renewcommand{\arraystretch}{1.3}
\begin{tabular}{r| llll | llll} \hline
&\multicolumn{4}{c|}{network Users} & \multicolumn{4}{c}{network URLs} \\
&AP & MF &EV&MP\textsuperscript{*}  &AP&MF&EV &MP\textsuperscript{*}  \\ \hline
$Q$ & 0.5254 & 0.3331 & 0.3390& \textbf{0.6008} & 0.2963& \textbf{0.5690}& 0.4530 & {0.3120} \\
\# & 239&100& 100 &\textbf{220} & 6870&100& 100 & \textbf{6653} \\
$C$ &0.7778 & 0.9209 & 0.8625 & \textbf{0.3037} & 0.4167 & 0.5962 &0.9233& \textbf{0.3657} \\
TPR ($\times 10^3$)& 1.49977 & 1.4282 & 0.3932 & \textbf{2.2193}& 95.678 & 28.54 &15.7567 & \textbf{106.59}\\
CR&12.8275 & \textbf{7.0166} &9.6902& 8.6286 & 272.5224 & 78.33 &\textbf{9.0630}&310.5238
\\ \hline
\multicolumn{9}{l}{\textsuperscript{*}\footnotesize{The results were achieved by setting $M$=0.2.
} }
\end{tabular}
\label{tab:Delicious2K}
\end{table*}

\noindent\textbf{Synthetic network II}

We further studied the advantage of the proposed method on synthetic networks with more variations.
The number of nodes in a homogeneous network was fixed to $100$, and the heterogeneous bicliques were generated in the same way as synthetic network I did. But the number of communities, denoted by $k$, within each homogeneous network were set different from $3$ to $5$.
Then, the node numbers of each community was generated via a $k$-dimensional symmetric Dirichlet distribution. To be specific,
a $k$-d vector was drawn according to the distribution parameterized with $\alpha=1$, whose summation is $1$.
Then it was multiplied by $100$ and outputted as the node numbers of communities after was rounded up/down to integers.
Moreover, the pair of inter- / intra- community link generating probabilities varies from $0.1/0.9$ to $0.9/0.1$. This variation leads to networks presenting characteristics with skewed degree distribution, decreasing community density and decreasing community separability.

Ten networks are generated for each pair of inter- /intra- community link generating probabilities, on which the community detecting task is conducted.
The average results over $10$ runs are shown in Figure \ref{fig:synII}. For both MF and EV, the performance was achieved with fixed $K=4$.

From the figures, several significant and interesting features can be observed.
First, the community structures discovered by AP show the worst quality, as from the lowest modularity values showing poor cluster property, the highest conductance values showing poor local community separability, and the highest cut ratio values corresponding to poor global separability. Comparing with the other three algorithms - MF, EV, and MP, such a incompetence is obviously due to lack of exploring heterogeneous information.
Second, the EV algorithm, at the beginning of its curve, presents the worst performance but maintains stable performance which is the second worst one throughout the x-axis. This is caused by that it only explores the heterogeneous information but ignores homogeneous information which sometimes plays more important role in discovering community structures than the former information.
Third, MF and the proposed MP achieve comparatively the best performance in different ways. Specifically, MF achieves the best performance in terms of the evaluation metrics, while MP achieves the best performance in terms of closeness to the ground-truth evaluation curve.


\begin{figure*}[ht]
\centering
\begin{subfigure}{1\textwidth}
\includegraphics[width=\textwidth]{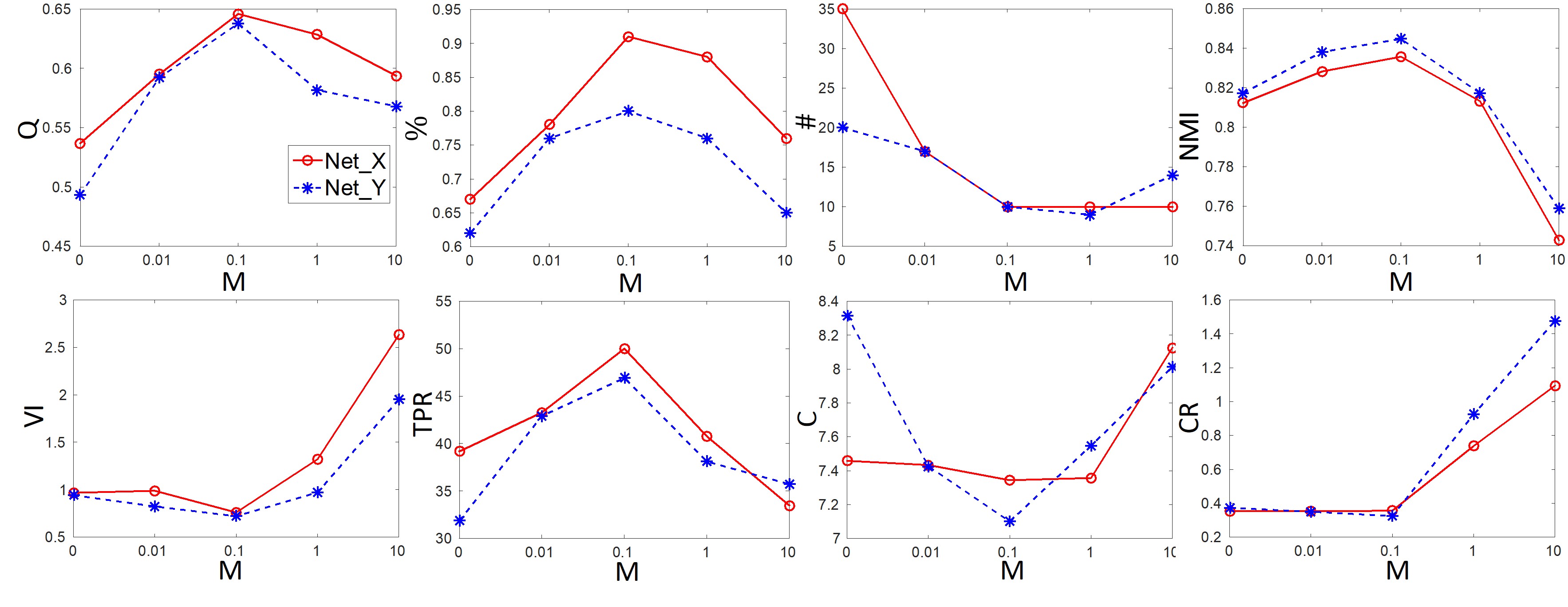}
\caption{Synthetic network }
\label{fig:synAccu}
\end{subfigure}
\\
\begin{subfigure}{1\textwidth}
\includegraphics[width=1\textwidth]{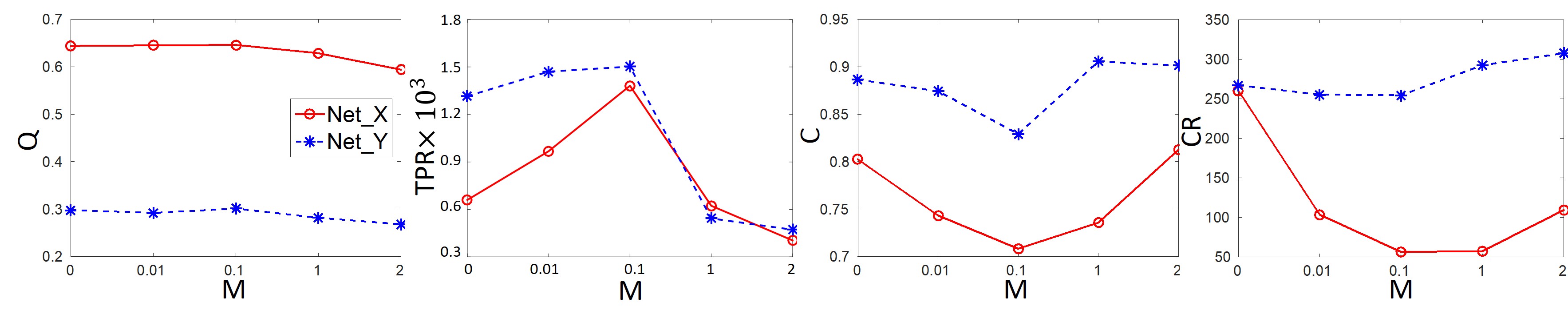}
\caption{DBLP}
\label{fig:synXAccu}
\end{subfigure}
\begin{subfigure}{1\textwidth}
\includegraphics[width=1\textwidth]{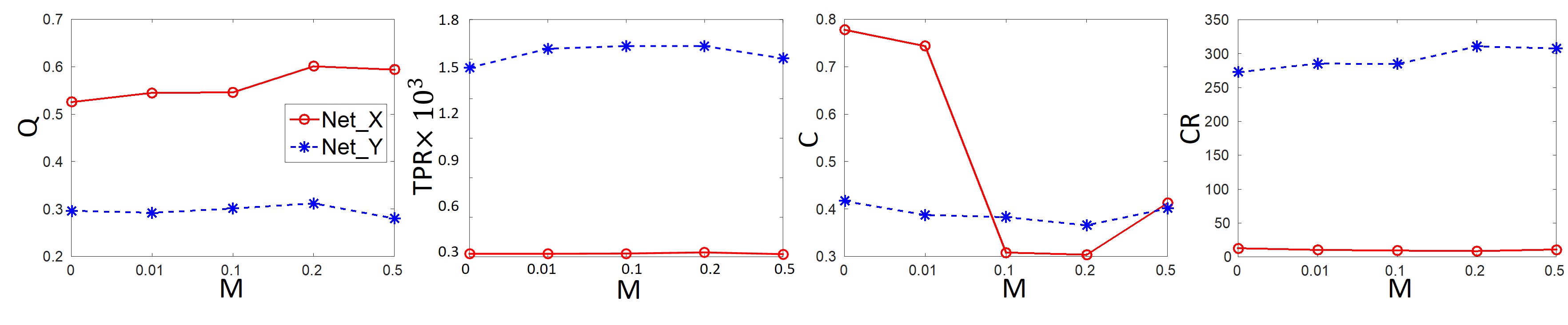}
\caption{Delicious-2K}
\label{fig:synXAccu}
\end{subfigure}
\caption{The effectiveness of the trade-off parameter $M$ for both synthetic and real-world networks.}
\label{fig:MResults}
\end{figure*}

\subsection{Real-world Datasets}

\noindent\textbf{DBLP}

The digital bibliography and library project dataset (DBLP) \footnote{http://dblp.uni-trier.de/db/} collected bibliographic records from major computer science journals and proceedings. Multi-typed objects such as authors, papers and text-rich data as well as their relationship were recorded.
A two-layered HMRNet was extracted from a subset of DBLP that was used and published in \cite{Hongbo2011DBLP} (Other subset sampling methods refer to \cite{leskovec2006sampling}).
The $X$-layer homogeneous network contains $28,702$ author nodes and $66,832$ co-author links, while the
$Y$-layer homogeneous network contains $28,569$ paper nodes and $216,713$ paper-relationship links.
There were $74,632$ paper-written-by-author heterogeneous single-relational links, from which $5,995$ heterogeneous bicliques were extracted.

Five evaluation metrics - Modularity (Q), number of communities, Conductance (C), TPR and CR - are applied, since no ground-truth community structure is acquired. The proposed algorithm is compared to one baseline algorithm AP and two state-of-the-art algorithms MF and EV. Since both MF and EV are unable to automatically determine the number of communities, it is fixed as $2000$ for MF. It is reasonable to set $2000$ for EV for fair comparison. However, due to computational infeasible, it is fixed as $100$ for EV. The results are summarized in Table \ref{tab:DBLP}.
It is easily seen that MP outperforms the baseline algorithm AP with a large margin, while achieves comparative performance with MF. On the other hand, EV, due to ignoring homogeneous information, only discovers community structures with extremely low density.

\noindent\textbf{Delicious-2K}

The Delicious social bookmarking dataset (Delicious-2K) is also usually used as benchmarks for community detection task\cite{CantadorRecSys2011}.
In our experiment, $1,861$ Users and $7,664$ friendship connections were extracted as the $X$-layer homogeneous network, while $69,226$ URLs and $653,386$ URL-to-URL links were extracted as the $Y$-layer homogeneous network. The URL-to-URL links were built with the similarities of URL titles.
For the heterogeneous information, there were $104,418$ User-to-URL bookmarks relationships, from which $5,100$ heterogeneous bicliques were constructed.

Like DBLP, five evaluation metrics are used to evaluate the quality of community structures detected by both the proposed algorithm and the state-of-the-art algorithms. The results are shown in Table \ref{tab:Delicious2K}.
As illustrated, the proposed algorithm achieves consistent performance as for DBLP dataset. Overall,
MP outperforms AP and EV, both of which consider partial relational information, while achieves comparatively results with MF.

\subsection{Study tradeoff parameter $M$}
Finally, how the performance of the proposed method is affected by the trade-off parameter $M$ is studied over both synthetic and real-world datasets.
The results are shown in Figure \ref{fig:MResults}.
 it is easily seen that a moderate $M$ achieves a good community detection results for both synthetic and real-world networks.



\section{Conclusions}
\label{sec:conclusion}
In this paper, we propose a framework for community detection in HMRNets. We designed an information exchange mechanism, which elegantly integrates the heterogeneous single-relational information into homogeneous single-relational information to enhance each homogeneous community detection results.
We formulated the community detection task in HMRNets with a maximization clustering objective, which is solved by an iterative message propagation procedure with the advantage that it consumes less memory and computing time.
Experimental results over both synthetic and real-world networks (DBLP and Delicious-2K) confirm the superiority of our proposed method.

\section*{Acknowledgments}
This work was partly supported by the National Natural Science Foundation of China under Grant 61702145, the Australian Research Council Projects under Grants FL-170100117 and DP-180103424.

\ifCLASSOPTIONcaptionsoff
  \newpage
\fi

\bibliographystyle{IEEEtran}
\bibliography{bibtex}

%
\begin{IEEEbiography}[{\includegraphics[width=1in,height=1.25in,clip,keepaspectratio]{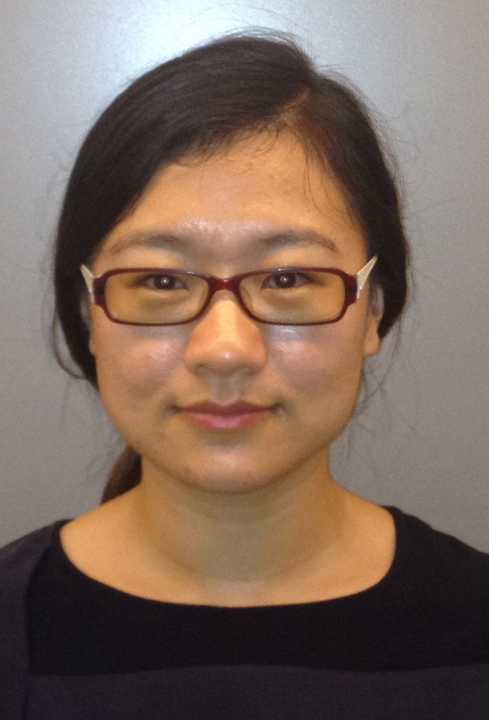}}]{Maoying Qiao}
received the B.Eng. degree in Information Science and Engineering from Central South University, Changsha, China, in 2009, and the M.Eng. degree in Computer Science from Shenzhen Institutes of Advanced Technology, Chinese Academy of Sciences, Shenzhen, China, in 2012, and the PhD degree in Computer Science in 2016 from the University of Technology Sydney.
Her research interests include machine learning and probabilistic graphical modeling.
\end{IEEEbiography}

\begin{IEEEbiography}[{\includegraphics[width=1in,height=1.25in,clip,keepaspectratio]{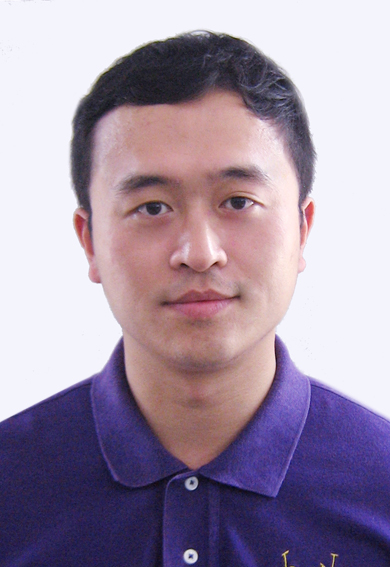}}]{Jun Yu}
(M'13) received his B.Eng. and Ph.D. from Zhejiang
University, Zhejiang, China. He is currently a
Professor with the School of Computer Science and
Technology, Hangzhou Dianzi University. He was an
Associate Professor with School of Information Science
and Technology, Xiamen University. From 2009 to 2011,
he worked in Singapore Nanyang Technological University.
From 2012 to 2013, he was a visiting researcher
in Microsoft Research Asia (MSRA). Over the past years,
his research interests include multimedia analysis,
machine learning and image processing. He has authored
and co-authored more than 50 scientific articles.
He has (co-)chaired for several special sessions, invited
sessions, and workshops. He served as a program committee member or reviewer
top conferences and prestigious journals. He is a Professional Member of the IEEE,
ACM and CCF.
\end{IEEEbiography}

\begin{IEEEbiography}[{\includegraphics[width=1in,height=1.25in,clip,keepaspectratio]{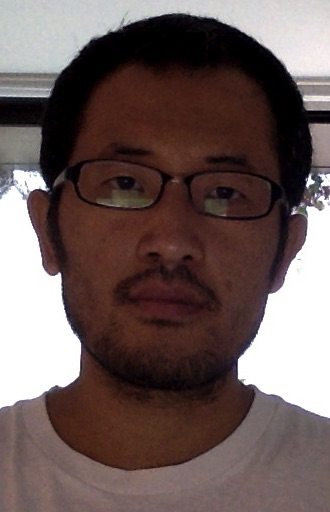}}]{Wei Bian}
(M'14) received the B.Eng. degree in electronic engineering and the B.Sc. degree in applied mathematics in 2005, the M.Eng. degree in electronic engineering in 2007, all from the Harbin institute of Technology, harbin, China, and the PhD degree in computer science in 2012 from the University of Technology, Sydney.
His research interests are pattern recognition and machine learning.
\end{IEEEbiography}

\begin{IEEEbiography}[{\includegraphics[width=1in,height=1.25in,clip,keepaspectratio]{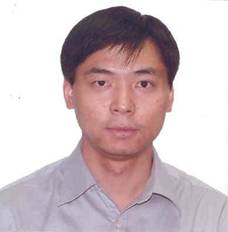}}]{Dacheng Tao}
Dacheng Tao (F'15) is Professor of Computer Science and ARC Laureate Fellow in the School of Computer Science and the Faculty of Engineering and Information Technologies, and the Inaugural Director of the UBTECH Sydney Artificial Intelligence Centre, at the University of Sydney. He mainly applies statistics and mathematics to Artificial Intelligence and Data Science. His research results have expounded in one monograph and 200+ publications at prestigious journals and prominent conferences, such as IEEE T-PAMI, T-IP, T-NNLS,T-CYB, IJCV, JMLR, NIPS, ICML, CVPR, ICCV, ECCV, ICDM; and ACM SIGKDD, with several best paper awards, such as the best theory/algorithm paper runner up award in IEEE ICDM��07, the best student paper award in IEEE ICDM��13, the 2014 ICDM 10-year highest-impact paper award, the 2017 IEEE Signal Processing Society Best Paper Award, and the distinguished paper award in the 2018 IJCAI. He received the 2015 Austrlian Scopus-Eureka Prize and the 2018 IEEE ICDM Research Contributions Award. He is a Fellow of the Australian Academy of Science, AAAS, IEEE, IAPR, OSA and SPIE.
\end{IEEEbiography}

\setcounter{equation}{0}
\pagebreak
\onecolumn
\begin{center}
\textbf{\large Supplemental Materials: Derivation for Messages}
\end{center}
The message sent from variable node $x_i$ to factor node $\delta_k$ consists of three parts: One is from $x_i$'s similarity factor, one is from all the other configuration constraint factor nodes' messages, one is from the biclique constraint factor nodes' messages. It is demonstrated in Figure 3a and the formula is:
\begin{align}
&\rho_{i \rightarrow \delta_k}(x_i) = 
s(i,x_i)+\sum \limits_{j:x_i\in C_j}\mu_{i \leftarrow \lambda_{C_j}}(x_i)+\sum_{k': k'\neq k} \alpha_{i \leftarrow \delta_{k'}}(x_i).
\notag
\end{align}
The message sent from factor node $\delta_k$ to variable node $x_i$ incorporates the summarization of the messages from all other variable nodes in the same layer and the potential of the factor node itself, which is clearly shown in Figure 3b and the max-sum formula is:
\begin{align}
&\alpha_{i \leftarrow \delta_k}(x_i) = 
\max_{x_{i':i' \neq i}}\left[ \delta_k(x_1,...,x_i,...x_M)+\sum_{i':i'\neq i} \rho_{{i'} \rightarrow \delta_k}(x_{i'}) \right]. \notag 
\end{align}
The message sent from variable node $x_i$ to factor node $\lambda_{C_j}$ is shown in
Figure 3c. Each single variable $x_i$ tries to convey all information (i.e. similarity, configuration constraints and other heterogenous biclique constraints) to factor node $\lambda_{C_j}$.
The specific formula is as below.
\begin{align}
&\gamma_{i \rightarrow \lambda_{C_j}}(x_i) = 
s(i,x_i) + \sum_{k'} \alpha_{i \leftarrow \delta_{k'}}(x_i)
+\sum_{\substack{j':j'\neq j,\\x_i \in C_{j'}}} \mu_{i \leftarrow \lambda_{C_{j'}}}(x_i).
\notag 
\end{align}
The message passing from factor node $\lambda_{C_j}$ to variable node $x_i \in C_j$ is visualized in Figure 3d. It collects information from all nodes in the biclique $C_j$, except $x_i$ itself, together with the factor potential $\lambda_{C_j}$ to output the message $\mu_{x_i \leftarrow \lambda_{C_j}}$ which is formulated as below.
\begin{equation}
\begin{aligned}
&\mu_{i \leftarrow \lambda_{C_j}}(x_i)=
\max \limits_{x_{i'}, i' \in \{C_j \backslash x_i\}}
\left[ \sum_{i'\in \{C_j\backslash x_i\}} \gamma_{{i'} \rightarrow \lambda_{C_j}}(x_{i'}) + \lambda_{{C_j}}(C_j) \right].
\end{aligned} \notag
\end{equation}
Finally, messages flowing into the variable node $x_i$ co-determine its labeling value (Figure 3e), which is formulated as:
\begin{align}
&\hat{x_i} = \arg\max_{x_i} 
s(i,x_i) +
\sum_{j:x_i\in C_j}\mu_{i \leftarrow \lambda_{C_j}} (x_i) + \sum_{j} \alpha_{i \leftarrow \delta_j} (x_i). \notag
\end{align}
Since the messages for network $Y$ from variable nodes to both biclique factor nodes and valid configuration factor nodes, as well as the messages from factor nodes to variable nodes, are in the same forms with messages for network $X$, here we do not list them in details.
\edef\mythmref{\ref{the:message}}
\label{appen:A}
\begin{proof}
\begingroup
\allowdisplaybreaks
\begin{equation}
\begin{aligned}
\alpha_{i \leftarrow \delta_k}(x_i) 
&= \max \limits_{x_{i':i' \neq i}}\left[ \delta_k(x_1,...,x_i,...x_M)+\sum \limits_{i':i'\neq i} \rho_{{i'} \rightarrow \delta_k}(x_{i'}) \right] \\
&=\begin{cases}
\sum \limits_{i':i'\neq i} \max \limits_{x_{i'}}~\rho_{{i'}\rightarrow \delta_k}(x_{i'}), ~~~~~~~~~~~~\mathrm{for}~x_i = k = i; \\
\sum \limits_{i':i'\neq i} \max \limits_{x_{i'}\neq k}~\rho_{{i'}\rightarrow \delta_k}(x_{i'}),~~~~~~~~~~~\mathrm{for}~x_i \neq k = i ;\\
\rho_{k\rightarrow k}(k) +\sum \limits_{i':i'\neq \{i,k\}} \max \limits_{x_{i'}}~\rho_{{i'}\rightarrow \delta_k}(x_{i'}), \\
~~~~~~~~~~~~~~~~~~~~~~~~~~~~~~~~~~~~~~~~\mathrm{for}~x_i = k \neq i;\\
\max
\begin{bmatrix}
\sum_{i':i'\neq i}~\max \limits_{x_{i'}\neq k}~\rho_{{i'}\rightarrow \delta_k}(x_{i'}), \\
\rho_{k\rightarrow k}(k) + \sum \limits_{i':i'\neq \{i,k\}}~\max \limits_{x_{i'}}~\rho_{{i'}\rightarrow \delta_k}(x_{i'})
\end{bmatrix} , \\ ~~~~~~~~~~~~~~~~~~~~~~~~~~~~
~~~~~~~~~~~\mathrm{for}~x_i \neq k \neq i.
\end{cases}
\end{aligned} \notag
\end{equation}
\endgroup
\begingroup
\allowdisplaybreaks
\begin{equation}
\begin{aligned}
 \mu_{i \leftarrow \lambda_{C_j}}(x_i) 
&
= \max \limits_{x_{i'},i' \in C_j^x}\left[ \sum \limits_{{i'}\in C_j^x} \gamma_{{i'} \rightarrow \lambda_{C_j}}(x_{i'}) + \lambda_{C_j}(C_j) \right] \\
& =\begin{cases}
\sum \limits_{{i'}\in C_j^x} \max \limits_{x_{i'}} \gamma_{i' \rightarrow \lambda_{C_j}} (x_{i'}),
\quad \mathrm{if}~x_i = x_{i'}^{\star} ~ \forall i' \in C_j^x,\\
~~~~~~~~~~~~~~~~~~~\mathrm{where}\quad x_{i'}^{\star} = \max \limits_{{i'} \in C_j^x}\gamma_{i' \rightarrow \lambda_{C_j}} (x_{i'}) ;\\
\max
\begin{bmatrix}
\sum \limits_{{i'}\in C_j^x} \max \limits_{x_{i'}} \gamma_{i' \rightarrow \lambda_{C_j}} (x_{i'}) - M, \\
~~ \sum \limits_{i' \in C_j^x} \gamma_{i' \rightarrow \lambda_{C_j}}(x_i)
\end{bmatrix}, \\
\qquad \qquad \qquad \qquad\qquad \qquad  \mathrm{if}~ x_i \neq x_{i'}^{\star}, \exists i' \in C_j^x.
\end{cases}
\end{aligned}
\label{eq:MuFractional}
\end{equation}
\endgroup
Replace
$\alpha_{i \leftarrow \delta_k}(x_i) = \tilde{\alpha}_{i \leftarrow \delta_k}(x_i) + \bar{\alpha}_{i \leftarrow \delta_k}$,
$\rho_{i \rightarrow \delta_k}(x_i) = \tilde{\rho}_{i \rightarrow \delta_k}(x_i) + \bar{\rho}_{i \rightarrow \delta_k}$
where
$\tilde{\alpha}_{i \leftarrow \delta_k}(x_i) =
\begin{cases} 0~ for~ x_i \neq k\\ \tilde{\alpha}_{i \leftarrow \delta_k}(k) ~ for~ x_i \neq k \end{cases}$
and $\bar{\alpha}_{i \leftarrow \delta_k} = \alpha_{i \leftarrow \delta_k}(x_i:x_i \neq k)$,
$\max \limits_{x_i} \tilde{\rho}_{i \rightarrow \delta_k}(x_i) = \max(0,\tilde{\rho}_{i \rightarrow \delta_k}(k))$ and
$\bar{\rho}_{i \rightarrow \delta_k} = \max \limits_{x_i, x_i \neq k}~\rho_{i \rightarrow \delta_k}(x_i)$.
The new messages are:
\begingroup
\allowdisplaybreaks
\begin{equation}
\begin{aligned}
&\rho_{i \rightarrow \delta_k}(x_i) = 
\begin{cases} s(i,x_i)+\sum\limits_{j:x_i\in C_j}\mu_{i \leftarrow \lambda_{C_j}}(x_i)+ \sum \limits_{k': k'\neq k} \bar{\alpha}_{i \leftarrow \delta_{k'}} ,\\
~~~~~~~~~~~~~~~~~~~~~~~~~~~~~~~~~~~~~~~~~~~~~~~\mathrm{for}~x_i = k;\\
s(i,x_i)+\sum\limits_{j:x_i\in C_j}\mu_{i \leftarrow \lambda_{C_j}}(x_i)+ \tilde{\alpha}_{i \leftarrow \delta_{k'}}(x_i= k') \\
~~~~~~~~~~
~+ \sum \limits_{k': k'\neq k} \bar{\alpha}_{i \leftarrow \delta_{k'}} ,~~~~~~~~~
~~~~~
~\mathrm{for}~x_i \neq k.
\end{cases}
\end{aligned}
\notag
\end{equation}
\endgroup
\begingroup
\allowdisplaybreaks
\begin{equation}
\begin{aligned}
&\alpha_{i \leftarrow \delta_k}(x_i) = 
\begin{cases}
\sum \limits_{i':i'\neq i} \max\{0,\tilde{\rho}_{{i'}\rightarrow \delta_k}(x_{i'}=k)\} + \sum \limits_{i':i'\neq i} \bar{\rho}_{{i'}\rightarrow \delta_k}, \\
~~~~~~~~~~~~~~~~~~~~~~~~~~~~~~~~~~~~~~~~~~~~~~~
~\mathrm{for}~x_i = k = i ;\\
\sum \limits_{i':i'\neq i} ~\bar{\rho}_{{i'}\rightarrow \delta_k},
~~~~~~~~~~~~~~~~~~~~~~~~~~~~~~~\mathrm{for}~x_i \neq k = i; \\
\tilde{\rho}_{k\rightarrow \delta_k}(k) +\sum \limits_{i':i'\neq \{i,k\}}
\max\{0,\tilde{\rho}_{{i'}\rightarrow \delta_k}(x_{i'} = k)\} \\
~~~~~~~~~~~
+ \sum \limits_{i':i'\neq i} \bar{\rho}_{{i'}\rightarrow \delta_k},~~
~~~~~~~~~~~~~~~\mathrm{for}~x_i = k \neq i;\\
\max \begin{bmatrix}
\sum \limits_{i':i'\neq i}~\bar{\rho}_{{i'}\rightarrow \delta_k},\\
\tilde{\rho}_{k\rightarrow \delta_k}(k) + \sum \limits_{i':i'\neq \{i,k\}}~\max\{0,\tilde{\rho}_{{i'}\rightarrow \delta_k}(x_{i'} = k)\} \\
~~~~~~~~~~~~~~~ +\sum \limits_{i':i'\neq i}~\bar{\rho}_{{i'}\rightarrow \delta_k}
\end{bmatrix}, \\
~~~~~~~~~~~~~~~~~~~~~~~~~~~~~~~~~~~~~~~~~~~~~~~~\mathrm{for}~x_i \neq k \neq i.
\end{cases}
\end{aligned}
\notag
\end{equation}
\endgroup
$\mu_{i \rightarrow \lambda_{C_j}}(x_i)$ is the same with (\ref{eq:MuFractional}).
\begingroup
\allowdisplaybreaks
\begin{equation}
\begin{aligned}
&\gamma_{i \rightarrow \lambda_{C_j}} (x_i = k')= s(i,x_i = k') + \tilde{\alpha}_{i \leftarrow \delta_{k'}}(x_i=k')
+\sum \limits_{k'=1}^M \bar{\alpha}_{i \leftarrow \delta_{k'}} + \sum_{\substack{j':j' \neq j,\\ x_i \in C_{j'}}} \mu_{i \leftarrow j'}(x_i).
\end{aligned} \notag
\end{equation}
\endgroup
Similar to the decomposition of the $\alpha$-s and $\rho$-s,
\begingroup
\allowdisplaybreaks
\begin{equation}
\begin{aligned}
\gamma_{i \rightarrow \lambda_{C_j}}(x_i) &= \tilde{\gamma}_{i \rightarrow \lambda_{C_j}}(x_i)+ \bar{\gamma}_{i \rightarrow \lambda_{C_j}},
\end{aligned} \notag
\end{equation} where
\endgroup
\begingroup
\allowdisplaybreaks
\begin{align}
&\tilde{\gamma}_{i \rightarrow \lambda_{C_j}}(x_i = k') = s(i,x_i = k') +
\tilde{\alpha}_{i \leftarrow \delta_{k'}}(x_i=k') 
+ \sum_{\substack{j':j'\neq j,\\x_i \in C_{j'}}}\mu_{i \leftarrow \lambda_{C_{j'}}}(x_i=k'), \notag 
\\
 &\bar{\gamma}_{i \rightarrow \lambda_{C_j}}(x_i)= \sum_{k'=1}^M \bar{\alpha}_{i \leftarrow \delta_{k'}}.
 \notag
\end{align}
\endgroup
Substitute the above equations into the formula $\mu$, and we obtain
\begingroup
\allowdisplaybreaks
\begin{equation}
\begin{aligned}
{\mu}_{i \leftarrow \lambda_{C_j}}(x_i) &= \tilde{\mu}_{i \leftarrow \lambda_{C_j}}(x_i) + \bar{\mu}_{i \leftarrow \lambda_{C_j}},
\end{aligned} \nonumber
\end{equation}
\endgroup
where
$\bar{\mu}_{i \leftarrow \lambda_{C_j}} = \bar{\gamma}_{i \rightarrow \lambda_{C_j}}
$ and
$\tilde{\mu}_{i \leftarrow \lambda_{C_j}}(x_i = k) $ is the same with (\ref{eq:MuFractional}), since the constant does not impact on the maximize.
{Solve for $\tilde{\alpha}_{i \leftarrow \delta_k} (x_i=k) = \alpha_{i \leftarrow \delta_k} (x_i = k) - \bar{\alpha}_{i \leftarrow \delta_k}$ and
$\tilde{\rho}_{i \rightarrow \delta_k}(x_i = k) = \rho_{i \rightarrow \delta_k}(x_i = k) - \bar{\rho}_{i \rightarrow \delta_k}$},
\begingroup
\allowdisplaybreaks
\begin{equation}
\begin{aligned}
\tilde{\rho}_{i \rightarrow \delta_k}(x_i = k) &= \rho_{i \rightarrow \delta_k}(x_i = k) - \bar{\rho}_{i \rightarrow \delta_k} \\
&= s(i,x_i=k)+\sum_{j:x_i\in C_j}\tilde{\mu}_{i \leftarrow \lambda_{C_j}}(x_i=k) + \sum_{j:x_i\in C_j}\bar{\mu}_{i \leftarrow \lambda_{C_j}} \\&
~~~~+ \sum_{k': k'\neq k} \bar{\alpha}_{i \leftarrow \delta_{k'}}~~ - \max\limits_{x_i, x_i \neq k}~\rho_{i \rightarrow \delta_k}(x_i) \\
&= s(i,x_i=k)+\sum_{j:x_i \in C_j}\{\tilde{\mu}_{i \leftarrow \lambda_{C_j}}(x_i=k) + \bar{\mu}_{i \leftarrow \lambda_{C_j}}\} \\
&+ \sum_{k': k'\neq k} \bar{\alpha}_{i \leftarrow \delta_{k'}}
- \max \limits_{x_i = k', x_i \neq k} \\
&~~~~
\begin{bmatrix}
s(i,x_i = k') + \sum_{j:x_i\in C_j}\{\tilde{\mu}_{i \leftarrow \lambda_{C_j}}(x_i=k') + \bar{\mu}_{i \leftarrow \lambda_{C_j}} \}\\
+ \tilde{\alpha}_{i \leftarrow \delta_{k'}}(x_i = k') + \sum\limits_{k':k' \neq k} \bar{\alpha}_{i \leftarrow \delta_{k'}}
\end{bmatrix} \\
&= s(i,x_i = k) + \sum_{j:x_i \in C_j} \tilde{\mu}_{i \leftarrow \lambda_{C_j}}(x_i = k)- \max \limits_{x_i = k', x_i \neq k} \\ &
\{s(i,x_i = k') + \sum_{j:x_i\in C_j} \tilde{\mu}_{i \leftarrow \lambda_{C_j}}(x_i = k')+ \tilde{\alpha}_{i \leftarrow \delta_{k'}}(x_i = k') \}.
\end{aligned}
\label{eq:finalRho}
\end{equation}
\endgroup
\begingroup
\allowdisplaybreaks
\begin{equation}
\begin{aligned}
\tilde{\alpha}_{i \leftarrow \delta_k} (x_i=k) &= \alpha_{i \leftarrow \delta_k} (x_i = k) - \bar{\alpha}_{i \leftarrow \delta_k} \\
&
= \alpha_{i \leftarrow \delta_k} (x_i = k) - \alpha_{i \leftarrow \delta_k}(x_i \neq k) \\
& = \begin{cases}
\sum \limits_{i':i'\neq i} \max\{0,\tilde{\rho}_{{i'}\rightarrow \delta_k}(x_{i'}=k)\} + \sum \limits_{i':i'\neq i} ~\bar{\rho}_{{i'}\rightarrow \delta_k} \\
~~~~~~~~
- \left[\sum\limits_{i':i'\neq i} \bar{\rho}_{{i'}\rightarrow \delta_k} \right],
~~~~~~~~~~~~~~~~~~~~~~\mathrm{for}~k = i; \\
\sum\limits_{i':i'\neq \{i,k\}}
\max\{0,\tilde{\rho}_{{i'}\rightarrow \delta_k}(x_{i'} = k)\} \\
+\tilde{\rho}_{k\rightarrow \delta_k}(k) +
\sum\limits_{i':i'\neq i} \bar{\rho}_{{i'}\rightarrow \delta_k}
- \max \\
\begin{bmatrix}
\sum\limits_{i':i'\neq i}~\bar{\rho}_{{i'}\rightarrow \delta_k},\\~~
\sum\limits_{i':i'\neq \{i,k\}}~\max\{0,\tilde{\rho}_{{i'}\rightarrow \delta_k}(x_{i'} = k)\} \\
+\tilde{\rho}_{k\rightarrow \delta_k}(k) + \sum\limits_{i':i'\neq i}~\bar{\rho}_{{i'}\rightarrow \delta_k}
\end{bmatrix},
~~~~\mathrm{for}~k \neq i;
\end{cases} \\
&=\begin{cases}
\sum\limits_{i':i'\neq i} \max\{0,\tilde{\rho}_{{i'}\rightarrow \delta_k}(x_{i'}=k)\},
~~~~~~~~~~~~~~~~~\mathrm{for}~k = i ;\\
\tilde{\rho}_{k\rightarrow \delta_k}(k) +\sum\limits_{i':i'\neq \{i,k\}}
\max\{0,\tilde{\rho}_{{i'}\rightarrow \delta_k}(x_{i'} = k)\} -\max \\
\begin{bmatrix} 0,\\
\tilde{\rho}_{k\rightarrow \delta_k}(k) + \sum_{i':i'\neq \{i,k\}}~\max\{0,\tilde{\rho}_{{i'}\rightarrow \delta_k}(x_{i'} = k)\}
\end{bmatrix}, \\
~~~~~~~~~~~~~~~~~~~~~~~~~~~~~~~~~~~~~~~~~~~~~~~~~~~~~
\mathrm{for}~k \neq i.
\end{cases}
\end{aligned}
\notag
\end{equation}
\endgroup
Finally the node labeling is computed as below.
\begingroup
\allowdisplaybreaks
\begin{align}
\hat{x_i} &=\arg\max_{x_i} ~p(x_i) \label{eq:pesudoMarginal} \\ %
&= \arg\max_{x_i} ~s(i,x_i) + \sum_{j:x_i \in C_j}{\mu}_{i \leftarrow \lambda_{C_j}}(x_i) +\sum_k \alpha_{i\leftarrow \delta_k}(x_i) \label{eq:decomposition}\\%
& =\arg\max_{x_i} ~s(i,x_i) +\sum_{j:x_i \in C_j}\{\tilde{\mu}_{i \leftarrow \lambda_{C_j}} (x_i) +\bar{\mu}_{i \leftarrow \lambda_{C_j}} \} 
+\sum_{k}\left(\tilde{\alpha}_{i \leftarrow \delta_k} (x_i)+ \bar{\alpha}_{i \leftarrow \delta_k} \right) \label{eq:noConstant}\\
&= \arg\max\limits_{x_i = k} ~s(i,x_i= k) +\sum_{j:x_i\in C_j}\tilde{\mu}_{i \leftarrow \lambda_{C_j}} (x_i=k) 
+ \tilde{\alpha}_{i \leftarrow \delta_k} (x_i=k) \label{eq:fromFinalRho} \\
& = \arg\max\limits_{x_i=k} ~s(i,x_i=k) +\sum_{j:x_i\in C_j}\tilde{\mu}_{i \leftarrow \lambda_{C_j}} (x_i=k) \label{eq:followFinalRho} \\
&~~~~~~~
+ \tilde{\alpha}_{i \leftarrow \delta_k}(x_i=k) -
\max \limits_{x_i = k', x_i \neq k}
\{s(i,x_i = k') \nonumber\\
&~~~~~~~+\sum_{j: x_i\in C_j}\tilde{\mu}_{i \leftarrow \lambda_{C_j}}(x_i = k')+ \tilde{\alpha}_{i \leftarrow \delta_{k'}}(x_i = k') \} \nonumber \\
& = \arg\max\limits_{x_i=k}~~\tilde{\rho}_{i \rightarrow \delta_k}(x_i = k) + \tilde{\alpha}_{i \leftarrow \delta_k}(x_i=k) \label{eq:finalMAX}.
\end{align}
\endgroup
Note that (\ref{eq:pesudoMarginal}) $\rightarrow$ (\ref{eq:decomposition}) is the pesudo marginal distribution from BP algorithm.
(\ref{eq:decomposition}) $\rightarrow$ (\ref{eq:noConstant}) and (\ref{eq:noConstant}) $\rightarrow$ (\ref{eq:fromFinalRho}) are the decompositions of the $\alpha$ and $\rho$.
(\ref{eq:fromFinalRho}) $\rightarrow$ (\ref{eq:followFinalRho}) is based on the truth that the constant will not affect the $\max$ result.
(\ref{eq:followFinalRho}) $\rightarrow$ (\ref{eq:finalMAX}) is the computation of $\tilde{\rho}_{i \rightarrow \delta_k}(x_i=k)$ in (\ref{eq:finalRho}).
\end{proof}


\end{document}